\pdfoutput=1
\documentclass{article}

\usepackage{arxiv}

\usepackage[utf8]{inputenc} 
\usepackage[T1]{fontenc}    
\usepackage{hyperref}       
\hypersetup{colorlinks,allcolors=black}
\usepackage{url}            
\usepackage{booktabs}       
\usepackage{amsfonts}       
\usepackage{nicefrac}       
\usepackage{microtype}      
\usepackage{lipsum}

\usepackage[ruled,vlined,linesnumbered]{algorithm2e}
\usepackage{graphicx}         
\usepackage{amsmath}
\usepackage{times}            
\usepackage{subcaption}
\title{Neuroevolution of Neural Network Architectures Using CoDeepNEAT and Keras}

\author{
  Jonas da Silveira Bohrer\thanks{\href{https://orcid.org/0000-0003-2868-3838}{https://orcid.org/0000-0003-2868-3838}}\\
  Institute of Informatics\\
  Federal University of Rio Grande do Sul\\
  Porto Alegre, Brazil\\
  \texttt{\href{mailto:jsbohrer@inf.ufrgs.br}{jsbohrer@inf.ufrgs.br}}
  \And
  Bruno Iochins Grisci\thanks{\href{https://orcid.org/0000-0003-4083-5881}{https://orcid.org/0000-0003-4083-5881}}\\
  Institute of Informatics\\
  Federal University of Rio Grande do Sul\\
  Porto Alegre, Brazil\\
  \texttt{\href{mailto:bigrisci@inf.ufrgs.br}{bigrisci@inf.ufrgs.br}}\\
  \And
  Marcio Dorn\thanks{\href{https://orcid.org/0000-0001-8534-3480}{https://orcid.org/0000-0001-8534-3480}}\\
  Institute of Informatics\\
  Federal University of Rio Grande do Sul\\
  Porto Alegre, Brazil\\
  \texttt{\href{mailto:mdorn@inf.ufrgs.br}{mdorn@inf.ufrgs.br}}
}

\begin{document}
\maketitle

\begin{abstract}
    Machine learning is a huge field of study in computer science and statistics dedicated to the execution of computational tasks through algorithms that do not require explicit instructions but instead rely on learning patterns from data samples to automate inferences. A large portion of the work involved in a machine learning project is to define the best type of algorithm to solve a given problem. Neural networks - especially deep neural networks - are the predominant type of solution in the field. However, the networks themselves can produce very different results according to the architectural choices made for them. Finding the optimal network topology and configurations for a given problem is a challenge that requires domain knowledge and testing efforts due to a large number of parameters that need to be considered. The purpose of this work is to propose an adapted implementation of a well-established evolutionary technique from the neuroevolution field that manages to automate the tasks of topology and hyperparameter selection. It uses a popular and accessible machine learning framework - Keras - as the back-end, presenting results and proposed changes concerning the original algorithm. The implementation is available at GitHub (\href{https://github.com/sbcblab/Keras-CoDeepNEAT}{https://github.com/sbcblab/Keras-CoDeepNEAT}) with documentation and examples to reproduce the experiments performed for this work.
\end{abstract}


\setlength\parindent{24pt}

\section{Introduction}

Evolutionary computation can be shortly described as the use of evolutionary systems as computational processes for solving complex problems \cite{DeJong:2016:ECU:3027779}. As discussed in \cite{DeJong:2016:ECU:3027779}, although one can trace its genealogical roots as far back as the 1930s, it was the emergence of relatively inexpensive digital computing technology in the 1960s that served as an essential catalyst for the field. The availability of this technology made it possible to use computer simulation as a tool for analyzing systems much more complicated than those analyzable mathematically.

Around the same time, machine learning emerged as a branch of AI proposing a more probabilistic approach to the search of artificial intelligence with systems that aimed to learn and improve without being explicitly programmed. Though it had an interesting premise, it only rose to its new level of popularity in the last few decades, justified by the increasing availability of large amounts of data and computing resources required for the extremely complex algorithms the field proposed.

These two fields, evolutionary computation and machine learning, come together in what is usually described as Evolutionary Machine Learning (EML), which presents hybrid approaches that use algorithms from one field in the search for better solutions in the other. These resulting approaches have been widely applied to real-world problems in various situations, including agriculture, manufacturing, power and energy, internet/wifi/networking, finance, and healthcare \cite{Al-Sahaf:2019}. 

Out of the many branches in EML, one of the most widely studied is Neuroevolution \cite{Floreano2008}, which is characterized by the act of building different aspects of neural networks through evolutionary algorithms (EAs) \cite{Back:1996:EAT:229867}. EAs are exceptionally well suited for this task because of their remarkable ability to find reasonable solutions in highly dimensional search spaces, such as exploring the multiple possibilities surrounding the definition of a neural network structure. Nonetheless, neuroevolution enables important capabilities that are typically unavailable to the more traditional gradient-based approaches like stochastic gradient descent (SGD) \cite{Rumelhart1986}, raising the level of automation beyond the initial perspective of only setting weights to preconfigured network topologies. These new capabilities include the search of ideal hyperparameters, structural parts, and even the rules for learning themselves \cite{Stanley2019}.

Of course, despite the significant benefits described of using neuroevolution, gradient-based methods still dominate many areas in machine learning where classification problems are easily differentiable with known topologies, as calculating weights through gradient descent methods is frequently more efficient than most evolutionary techniques. Still, neuroevolution finds its niches in domains where ideal topologies are yet to be discovered, such as in the meta-learning field \cite{Pappa2014} and the reinforcement learning field \cite{gascompetitivereinf}, proving to be a scalable option in these domains \cite{salimans2017evolution}.

A great example of the early neuroevolution approach successfully applied to a wide range of problems is the NeuroEvolution of Augmenting Topologies (NEAT) algorithm \cite{NEAT}, which is the starting point of this work. NEAT's main idea was to generate neural networks by associating similar parts of different neural networks through mutations (adding or removing nodes and connections) and crossovers (swapping nodes and connections) with a historical markings mechanism that simplified the identification of network similarities. Most importantly, it managed to implement a remarkable diversity preservation mechanism (named speciation), enabling the evolution of increasingly complex topologies by allowing organisms to compete primarily within their niches instead of with the population at large.
    
But NEAT did not age well throughout the last decade, despite its remarkable success in multiple use cases \cite{Stanley2019} - like the notorious discovery through NEAT of the most accurate measurement yet of the mass of the top quark, which was achieved at the Tevatron particle collider \cite{PhysRevLett.102.152001} - where the minimal structure was a lot more of a priority. Compared to state of the art in modern deep learning research, the networks generated by the original NEAT algorithm are easily surpassed in dimension and consequently effectiveness when compared to networks used in currently widespread problems like image recognition or text recognition. In these problems, thousands of nodes and hundreds of thousands to millions of connections are necessary to process information about complex data sources accordingly. The growth tendency in network dimensions comes directly from the availability of unprecedentedly cheap and powerful computing resources and large datasets, as seen in the latest years. This availability is not only reducing the need for minimal structures in standard neural networks but also resulting in the perfect conditions for the practical usage and consequent popularization of all sorts of creative solutions involving different approaches to the traditional neural network topology, such as deep networks, convolutional networks, LSTM networks, graph networks, relational networks and more, contributing to yet one more weakness in standard NEAT.

This rapid popularization of different types of neural networks brought into the neuroevolution field challenges to try new techniques by combining and expanding these varied components into appropriate topologies and configurations to solve problems even more effectively and is also referred to as the neural architecture search problem \cite{DBLP:journals/corr/ZophL16}. Adaptations in the traditional neuroevolution algorithms to face this evolving environment of possibilities and need for larger structures are popular at the moment \cite{Stanley2019}.

These adaptations can be seen in successful recent approaches like network generation and feature selection in highly dimensional datasets \cite{8790302}, applications to the identification of gene expression patterns in cancer research \cite{pmid30521855}, reinforcement learning tasks \cite{salimans2017evolution}, learning policies for data augmentation \cite{cubuk2018autoaugment}. There were also multiple successors of NEAT throughout the years, like the notorious HyperNEAT \cite{hyperneat}, DeepNEAT, and CoDeepNEAT variations \cite{DBLP:journals/corr/MiikkulainenLMR17}, which are the focus of this work.

    \subsection{Motivation}
        Although there are implementations of algorithms like NEAT and its variations from their authors, they consist of self-contained code that can be expanded but presents barriers in terms of directly connecting to other popular Machine Learning frameworks that researchers, students, or scientists are more likely to be familiar with. Keras \cite{chollet2015keras}, TensorFlow \cite{tensorflow2015-whitepaper}, PyTorch \cite{paszke2017automatic} and other similar frameworks contain several functionalities that may come in handy when developing or analyzing machine learning models, which is a key element in validating the resulting models from neural architecture search algorithms in practical scenarios.
        
        As of the moment of this work, both NEAT and HyperNEAT have been explored in public implementations \footnote{\href{https://github.com/crisbodnar/TensorFlow-NEAT}{https://github.com/crisbodnar/TensorFlow-NEAT}}\footnote{\href{https://pypi.org/project/neat-python/}{https://pypi.org/project/neat-python/}} using these frameworks but few or lacking implementations of CoDeepNEAT have been found, presenting a direct opportunity to bring this method to a more accessible context. On the other hand, it is unclear from the original work \cite{DBLP:journals/corr/MiikkulainenLMR17} whether the algorithm is suitable for practical applications and can be used in simple hardware environments or not. Verifying these aspects allows us to identify possible improvements to the base algorithm, such as different crossover operations,  or mutation operations. Additionally, having an implementation based on a broad framework facilitates these experiments for the overall scientific community.
        
        With these aspects in mind, this work established the implementation of an algorithm based on CoDeepNEAT in an accessible and popular framework and adapted based on different approaches seen in the literature. The objective is to validate the complexity of the process of implementing such an algorithm and verifying in practice if this type of solution is useful without massive hardware requirements. The framework of choice for this implementation is Keras, a user-friendly and high-level Python package for machine learning development and management, as opposed to the low-level and complex usability found in other popular options like directly using TensorFlow or PyTorch, for instance. Still, the back-end used for Keras is TensorFlow.
        
        \subsection{Proposed methodology}
        
        The implementation requires the following fundamental working parts before initial testing:
            \begin{itemize}
                \item Genetic algorithm structure (to support iterations).
                \item Graph generation structure (to generate graphs for modules and blueprints).
                \item Module population management structure (to generate modules, manage speciation, fitness sharing).
                \item Blueprint population management structure (to generate blueprints, manage speciation, assembling, training, fitness evaluation, and fitness sharing).
                \item Similarity metric and clusterization technique used for speciation (to compare individuals).
                \item Crossover technique used for reproduction (to evolve individuals through sexual reproduction).
                \item Mutations (to evolve individuals through asexual reproduction).
                \item Logging structure (to follow up the iteration process).
            \end{itemize}
            
        Additional modifications can be explored, such as:
            \begin{itemize}
                \item Alternative crossover operations.
                \item Alternative mutation techniques.
                \item Alternative similarity metrics.
            \end{itemize}
        
        Once the implementation was defined, initial experimentation used the MNIST \cite{MNIST} dataset. Final experiments were executed using the CIFAR-10 \cite{CIFAR-10} dataset as done in the original paper to compare results and discuss the amount of time and computing power required for this approach considering academic use. Preliminary tests point that the required time for complete runs of the implementation using these datasets varies around 6 and 12 hours, considering limited hardware configurations and reduced parameters, which will be described in the process. Most of the computation necessary is dedicated to training the networks for fitness evaluation during evolution.
        
        Section \ref{chap:algoschapter} explores the required algorithms and concepts to develop the proposed work. Section \ref{chap:implementation} details the implementation and defines the usage of the concepts described in Section \ref{chap:algoschapter}, while Section \ref{chap:experiments} describes the experiments performed using the implementation and discuss the results, comparing them to the original CoDeepNEAT experiments and highlighting possible improvements.

\section{Computational methods and concepts} \label{chap:algoschapter}

    This Section briefly describes the most important algorithms and concepts related to the execution of this work and similar works in the EML field. Most recurrent terms are explained here and referenced in the next sections.
    
    \subsection{Genetic algorithms} \label{gasection}

        Genetic algorithms (GAs) are computational methods whose fundamental principle is the evolution of candidate solutions over iterations. Strongly based on behaviors of populations of biological organisms, they represent a predominant type of evolutionary algorithm (EA) in the evolutionary computation field, having been applied for decades in the solution of optimization problems since their first concrete description by J.H. Holland \cite{Holland:1992:ANA:531075}.
        
        As described in \cite{Beasley93anoverview}, in nature, individuals in a population compete with each other for resources such as food, water, and shelter. Also, members of the same species often compete to attract a mate. Those individuals who are most successful in surviving and attracting mates will have relatively larger numbers of offspring. Poorly performing individuals will produce few or even no offspring at all. This means that the genes from the highly adapted or "fit"  individuals will spread to an increasing number of individuals in each successive generation. The combination of good characteristics from different ancestors can sometimes produce "superfit"  offspring whose fitness is higher than that of either parent. In this way, species evolve to become more and more well suited to their environment.
        
        Adapting these concepts into a generalistic environment, standard GAs work with populations of "individuals" that represent solutions to a given problem. During multiple generations, these individuals are evaluated by a fitness function and are assigned a score. The scores are then used to decide what are the most "fit"  individuals for reproduction or survival. Through reproduction, "offspring"  is generated by combining the "genetic" features of their parents, occasionally generating better scoring solutions in the process. Individuals that are not fit enough for reproduction usually represent "bad" solutions, being less favored during the reproduction process and commonly replaced by new individuals.
        
        With the iteration of generations, genetic features that produce good solutions are likely to spread across the population of individuals, being passed on to offspring generated through reproduction or only by preservation mechanisms such as elitism, which consists in preserving a portion of the best scoring solutions over generations. GAs tend to converge over generations to optimum solutions, but require attention to issues such as keeping diversity (or, in other words, avoiding solutions to become extremely similar genetic representations). To address these matters, additional techniques can be implemented, such as preserving groups of similar solutions as "species," or including "mutations"  by altering genetic features in a defined fashion and consequently introducing changes to the populations. Pseudocode representing a standard procedure for GAs based on \cite{l.davis1991handbook-of-gen} can be seen in Algorithm \ref{alg_ga}, employing fitness evaluation, elitism, crossovers, and mutations to individuals over generations.
        
        \begin{algorithm}[ht]
        \caption{Basic genetic algorithm structure} \label{alg_ga}
        \SetAlgoLined
        \KwData{$N$: number of generations, $S$: population size, $E$: elitism rate, $C$: crossover rate, $M$: mutation rate}
        \KwResult{evolved candidate solutions}
        \Begin{
        initialize population with $S$ individuals\;
        \For{individual in population}{
            evaluate fitness\;
        }
            \For{generation in $N$}{
                apply elitism to $E\times S$ most fit individuals\;
                apply crossover to $C\times S$ most fit individuals\;
                apply mutations to $M\times S$ random individuals\;
                \For{individual in population}{
                    evaluate fitness\;
                }
            }
         }
        \end{algorithm}
        
        Being a trendy algorithm branch, GAs have evolved into different approaches. They have been successfully applied to a wide variety of optimization problems, such as protein folding \cite{Unger:1993:GAP:645513.657747}, selection of subsets of features to represent classification patterns \cite{Yang1998}, optimum container placement in container loading problems \cite{BORTFELDT2001143}, optimization of bank lending decisions \cite{METAWA201775}, increasing the longevity of wireless sensor networks \cite{Yuan2017} or approximating the mass of the top quark, which was achieved at the Tevatron particle collider through NEAT \cite{PhysRevLett.102.152001}.
        
    \subsection{Clustering algorithms} \label{clusteringsection}
        
        Clustering algorithms are a class of methods that focus on grouping or classifying representations of data in a common environment to sets of members called "clusters." They are well suited for data domains with no pre-defined classes, generating classifications based on custom metrics that evaluate the distance or similarity between the data samples.
        
        The specific clustering method used in this study is K-means \cite{Lloyd:2006:LSQ:2263356.2269955}, a popular partitioning algorithm based on specifying an initial quantity of groups, and iteratively reallocating objects among these groups until convergence. The algorithm assigns each data vector to the cluster whose center (also called "centroid") is nearest to the sample in the dimensional space that represents the data. The center is the average of all the points in the cluster, and its coordinates are the arithmetic mean for each dimension separately over all the points in the cluster \cite{DBLP:journals/corr/abs-1205-1117}. Algorithm \ref{alg_kmeans} adapted from \cite{DBLP:journals/corr/abs-1205-1117} describes the standard procedure for K-means. 
        
        \begin{algorithm}[ht]
            \caption{K-means algorithm} \label{alg_kmeans}
            \SetAlgoLined
            \KwData{$K$: number of clusters, $samples$: vectors representing the data, $tolerance$: minimum improvement rate to continue processing}
            \KwResult{Clusterization of the vectors}
            \Begin{
                Initialize the vectors of the $K$ $clusters$ (randomly, for instance)\;
                \While{not converged according to $tolerance$}{ \label{whilestep}
                    \For{every sample vector in samples}{
                        Compute the distance between the $sample$ vector and every $cluster$'s vector\;
                        Re-compute the closest vector to the $sample$ vector, using a learning rate that decreases in time\;
                    }
                }
                Return the clusterization\;
            }
        \end{algorithm}
        
        After clusters are defined, new samples of data can be integrated without the need for recreating the clusters. This can be done simply by using the \textit{nearest centroid} method, which is the execution of Algorithm \ref{alg_kmeans} from the $while$ step in line \ref{whilestep}, without initializing the $K$ clusters \cite{hastie_09_elements-of.statistical-learning}.
        
        In GAs, clustering algorithms can be applied to generate species or groups that share similar genetic information in the population. The species can be used in multiple strategies such as increasing population sizes without increasing the amount of fitness evaluations \cite{934284}, by evaluating one representative of the species at a time or ensuring diversity by preserving different groups of solutions, as in the case of NEAT \cite{NEAT}.
        
    \subsection{Artificial neural networks}
    
        Artificial neural networks (or directly "neural networks") are machine learning models composed of multiple information-processing units called "neurons," which are connected in different fashions to represent and approximate mathematical functions. Based on human biology, these models aspired to mimic the capability of the human brain to organize its structural constituents, known as neurons, to perform certain computations (e.g., pattern recognition, perception, and motor control) many times faster than the fastest digital computer in existence today \cite{haykin2009neural}.
        
\begin{figure}[ht]
            \centering
                \includegraphics[width=0.7\textwidth]{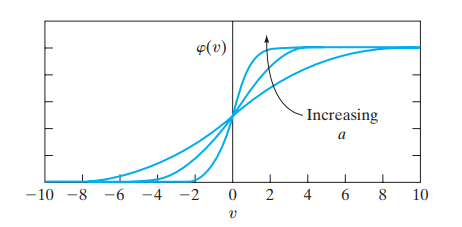}
                \caption{\textbf{A sigmoid function for varying slope parameter $a$}. Source: \cite{haykin2009neural}.}
                \label{fig:sigmoid}
        \end{figure}

        In practice, the neurons (also called nodes) that constitute neural networks are simple representations of mathematical functions that process the inputs they receive and output a value. They are composed of three main parts:
        \begin{itemize}
            \item A set of synaptic weights, each representing a value to be multiplied by the input signal of each connection they are assigned to.
            \item Summing junction, usually a linear combiner that sums the weighted input signals from the connections.
            \item Activation function, which models the output signal of the neuron to a defined amplitude. One of the most common activation functions, the \textit{sigmoid} function, is represented in Figure \ref{fig:sigmoid}.
            
        \end{itemize}
        
        \begin{figure}[ht]
            \centering
                \includegraphics[width=0.7\textwidth]{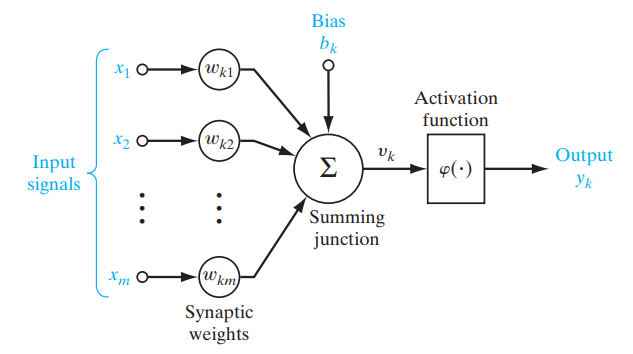}
                \caption{\textbf{A visualization of the components of a neural network.} Source: \cite{haykin2009neural}.}
                \label{fig:nn_neuron}
        \end{figure}
        
        In addition to the described components, Figure \ref{fig:nn_neuron} also displays the presence of a bias factor. The bias has the effect of increasing or lowering the net input of the activation function, depending on whether it is positive or negative, respectively \cite{haykin2009neural}. 
        
        In mathematical terms, the neuron $k$ depicted in Figure \ref{fig:nn_neuron} is described in \cite{haykin2009neural} by the three equations:
        
        \begin{equation}
            u_{k} = \sum_{j=1}^{m} w_{kj}x_{j}
        \end{equation}
        
        \noindent where $x_{1}$, $x_{2}$, ..., $x_{m}$ are the input signals, $w_{k1}$, $w_{k2}$, ..., $w_{km}$ are the respective synaptic weights of neuron $k$, $u_{k}$ is the linear combiner output due to the input signals;
        
        \begin{equation}
            v_{k} = (u_{k} + b_{k})
        \end{equation}
        
        \noindent where $b_{k}$ is the bias and $v_{k}$ is the resulting value from the summing junction after including the bias to $u_{k}$; and

        \begin{equation}
            y_{k} = \phi (v_{k})
        \end{equation}
        
        \noindent where $\phi$ is the activation function, and $y_{k}$ is the output signal of the neuron. One example of an activation function is depicted in Figure \ref{fig:sigmoid}, where a \textit{sigmoid} function is applied to an output signal generating a new output signal contained inside a restricted amplitude.
        
        The interconnections of the signals inside neurons and between neurons can be easily represented as signal-flow graphs \cite{4051460}, where the neurons are usually defined as "nodes." The connections can take multiple forms and are ruled by the synaptic weights. The synaptic weights that regulate these connections are subject to adjustments through procedures called "learning algorithms" and represent the knowledge acquired during the learning process. The learning algorithms are constituted frequently by the act of exposing the model to data samples and modifying the synaptic weights as the model "learns"  the patterns of the data. This procedure of exposing the model to data and evaluating its response is called supervised learning, the most common approach to "training"  neural networks to date.
        
\begin{figure}[ht]
            \begin{minipage}[t]{.39\textwidth}
                \begin{subfigure}[t]{\linewidth}
                    \caption{}
                    \vspace{.14\textwidth}
                    \includegraphics[width=5.5cm]{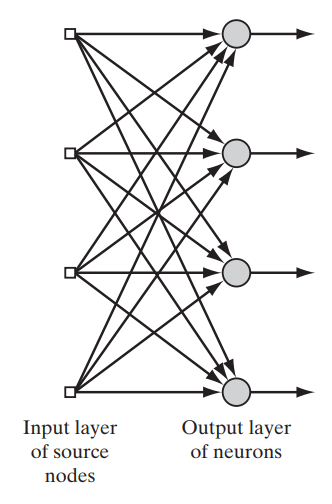}
                    \label{fig:feedforward}
                \end{subfigure}
            \end{minipage}
            \begin{minipage}[t]{.59\textwidth}
                \begin{subfigure}[t]{\linewidth}
                    \caption{}
                    \centering
                    \includegraphics[width=7.5cm]{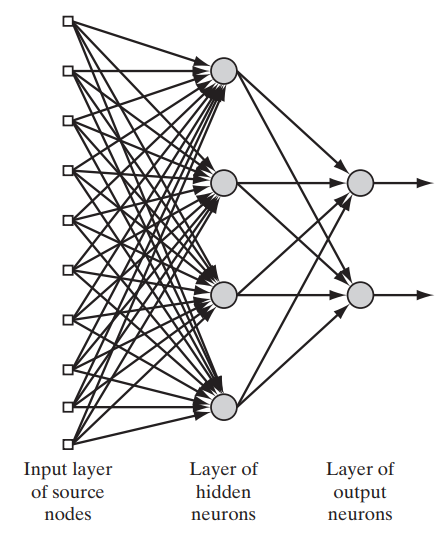}
                    \label{fig:multilayer_feedforward}
                \end{subfigure}
            \end{minipage}
            \caption{\textbf{Feedforward network structures.} A classic feedforward network structure (a) and a classic multilayer feedforward network structure (b). Source: \cite{haykin2009neural}.}
            \label{fig:mlp}
        \end{figure}

        Network architectures resulting from the interconnection of nodes can be classified in multiple definitions, but the most important initial taxonomies for this study are the {feedforward networks} and the {multilayer feedforward networks}. {Feedforward networks} are simply networks organized in a way that input nodes directly connect to output nodes to produce output signals, as in Figure \ref{fig:feedforward}. Multilayered feedforward networks implement the same logic but include nodes in divisions called "layers," representing sets of nodes that connect to other layers. Intermediate layers are commonly addressed as "hidden layers." An example can be seen in Figure \ref{fig:multilayer_feedforward}, where the input nodes connect to an intermediate layer, which connects to the output layer.
        
        From this initial notion of stacking layers was created, for example, the currently prevalent \textit{deep learning} branch \cite{deng2014deep} in the machine learning field. Deep learning architectures are commonly characterized by the connection of multiple layers of neurons in neural networks that take profit from extracting different levels of patterns in the input data with each layer. These architectures have been applied to fields including speech recognition \cite{pmlr-v48-amodei16}, image classification \cite{Rawat:2017:DCN:3146084.3146086}, natural language processing \cite{Goldberg:2017:NNM:3110856}, medical image analysis \cite{LitjensKBSCGLGS17} and more, in some cases reaching levels of confidence superior to those of human experts \cite{6248110}.
        
    \subsection{Neuroevolution algorithms} \label{neatsection}
        
        Neuroevolution is a field of study dedicated to the generation and improvement of neural networks using evolutionary algorithms (EAs). Traditionally associated with the generation of neuron weights through evolution, current approaches associated with the field focus on multiple aspects of the construction of a network, such as learning their building blocks (activation functions), hyperparameters (learning rates), architectures (number of neurons per layer, number of layers, and which layers connect to which) and even the rules for learning themselves \cite{Stanley2019}.
        
        One famous neuroevolution approach called Neuroevolution of Augmenting Topologies \cite{NEAT} and some of its variations will be explored in the next subsections and exemplify some use cases that benefit from the capacity of EAs to find adequate solutions for very complex problems, like the weight search, topology search, and hyperparameter search topics for neural networks.
    
    \subsubsection{NEAT}
    
        Neuroevolution of Augmenting Topologies \cite{NEAT}, also called NEAT, is an algorithm designed for neural network topology construction. NEAT uses a genetic algorithm structure to generate small initial networks that evolve and grow over generations by adding neurons and connections and adjusting their weights to generate structures capable of performing well while keeping them minimal in size. This minimalist aspect of NEAT is one of its core differences to other neuroevolution algorithms, as it focuses on only adding neurons or connections when they have an active impact in the network's performance \cite{NEAT}.
        
        Starting with an initial population of small networks based on a common topology, NEAT evaluates changes to these networks iteratively by adding and removing neurons and connections across generations. The algorithm defines the genome that describes the nodes and connections by a mechanism called genetic encoding, used in the operations that modify the network structures through classic genetic algorithm operators such as crossover and mutations.
        
\begin{figure}[ht]
            \centering
            \includegraphics[width=0.5\textwidth]{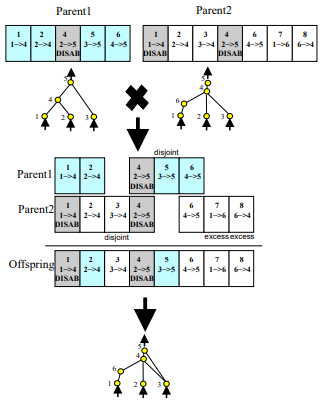}
            \caption{\textbf{NEAT crossover operation example.} Although Parent 1 and Parent 2 look different, their historical markings (shown at the top of each gene) tell us which genes match up with which. Even without any topological analysis, a new structure that combines the overlapping parts of the two parents, as well as their different parts, can be created. Matching genes are inherited randomly, whereas disjoint genes (those that do not match in the middle) and excess genes (those that do not match in the end) are inherited from the more fit parent. Source: \cite{NEAT}.}
            \label{fig:neat_crossover}
        \end{figure}

        Mutation operators in NEAT work by adding nodes and connections or by disabling existing connections, avoiding changes that affect the functionality of the network. Crossover operators, on the other hand, are a much more complicated operation as it requires vast exchanges of genetic information that may cause resulting networks not to work correctly. To solve this, NEAT implements a historical markings mechanism, identifying nodes and connections with numerical identifiers. Parts of networks that share the same origin will share the same identifiers. Thus the algorithm can recognize common structures in a simple way and exchange genetic information without generating defective networks (Figure \ref{fig:neat_crossover}).
        
        Before applying crossover operations between networks, NEAT must ensure that the chosen networks are compatible to a certain degree. The algorithm manages this situation by applying a speciation technique to the population of solutions, dividing it in different species generated by similarity, allowing organisms to compete primarily within their niches instead of with the population at large. With this factor, different network topologies have a chance to evolve at their own pace instead of being instantly replaced by fast-converging networks that achieve better results in early generations.
        
        After its conception, NEAT was used in a wide variety of use cases, especially in settings where small networks were required because of performance constraints, such as in robotics \cite{dsilva:ms06}, physics \cite{PhysRevLett.102.152001}, content generation for video games \cite{Hastings:2010:GAR:1810136.1810137} and more, as well as inspiring multiple variations of its core ideas, as explored in the next subsections.

    \subsubsection{HyperNEAT}
        HyperNEAT or hypercube-based NEAT \cite{hyperneat} is probably the major extension of NEAT to date and has become a complex topic on its own, inspiring multiple approaches based on its success. Using connective CPNNs (Compositional Pattern Producing Networks) to represent connectivity patterns as functions of the Cartesian space \cite{hyperneat}, HyperNEAT exploits regularities in the data domain to evolve larger neural networks. In other words, the use of CPNNs enables indirect encoding, a principle based on attributing the discovery of patterns and regularities to the algorithm itself, relying as little as possible on direct encoding from designers.
        
\begin{figure}[ht!]
        \centering
            \includegraphics[width=1.\textwidth]{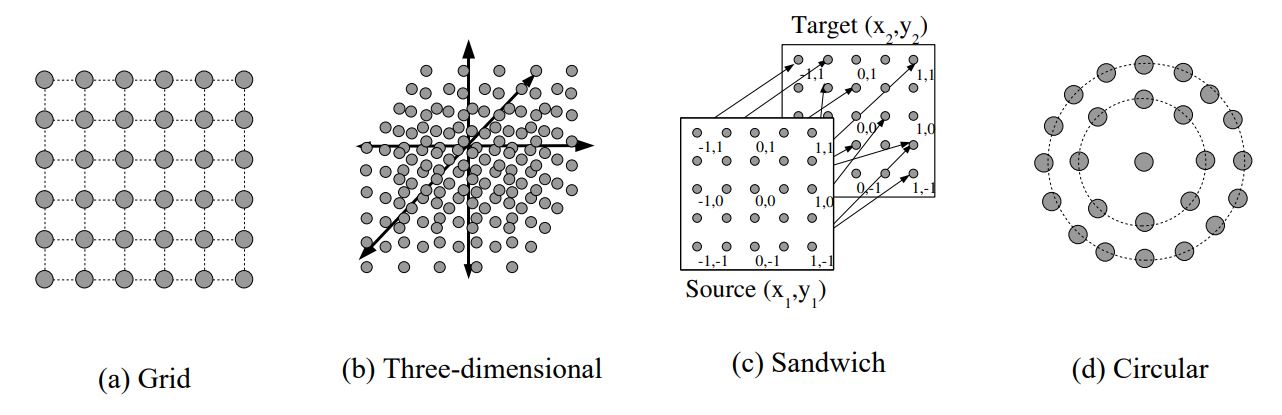}
           \caption{\textbf{Examples of configurations obtained using HyperNEAT.} This figure shows (a) a traditional 2D substrate of connections, (b) a three-dimensional configuration of nodes, (c) a “state-space sandwich” configuration in which a source sheet of neurons connects directly to a target sheet, and (d) a circular configuration. Different configurations are likely suited to problems with different geometric properties. Source: \cite{hyperneat}.}
            \label{fig:hyperneat_substrates}
        \end{figure}

        Moreover, indirect encoding aims to access regularities not commonly addressed by conventional neural network learning algorithms, being capable of inferring constructions like, for instance, convolution. Examples of node configurations obtained using HyperNEAT are shown in Figure \ref{fig:hyperneat_substrates}. 
        
        HyperNEAT also means a breakthrough from NEAT by allowing the evolution of much larger neural networks than the previous algorithm. By abstracting the mapping of spatial patterns generated by small CPNNs into connectivity patterns, HyperNEAT allows the generated networks to be scaled in a customizable manner (up to millions of connections, for instance). It can better adapt to more complex applications such as evolving controller parts of legged robots \cite{4983289}, learning to play Atari games \cite{hausknecht:tciaig13}, combining SGD and indirect encoding for network evolution \cite{Fernando:2016:CED:2908812.2908890} and even directly evolving modularity of components \cite{Verbancsics:2011:CCE:2001576.2001776}.

    \subsubsection{DeepNEAT and CoDeepNEAT} \label{codeepneatsection}
        Alternatively, a more recent path taken from NEAT was the DeepNEAT variation and, subsequently, the CoDeepNEAT variation \cite{DBLP:journals/corr/MiikkulainenLMR17}. Both cases, which are very tied, differ from HyperNEAT in that they do not aim to learn connectivity from geometric regularities in the data, but instead in assembling nodes based more directly in adaptations of the fitness evaluation process of NEAT.
        
        DeepNEAT can be summarized as an extension of NEAT that considers entire layers as genes instead of considering single neurons when forming structures. The focus now is to define compositions of layers instead of picking neurons and their connections one by one, generating larger and deeper networks suited to solving larger-scale problems than the ones NEAT was meant to solve in the past, while not minding the indirect encoding factor of HyperNEAT and considering pre-established components like different types of layers.
        
        Similarly to the original NEAT algorithm, DeepNEAT follows a standard genetic algorithm structure to find its solutions: it starts by creating an initial population of individuals, each represented by a graph, and evolves them over generations. During these generations, the individuals are recreated by adding or removing structural parts (nodes and edges) from their graphs through mutation, while keeping track of changes through a historical markings mechanism. Using the historical markings, chromosomes are compared in every generation using a similarity metric, being classified into subpopulations called species. Each species is evaluated by the shared fitness of its individuals, calculated by a fitness sharing function. This shared score is used to evaluate the quality of the species in each generation. Finally, the surviving species evolve separately from each other through crossovers (exchanging genetic information) among its constituent individuals, and the next generation takes place.
        
        The changes to the main algorithm of NEAT in how nodes now represent layers imply additional aspects that must be considered when defining a layer in DeepNEAT: what is the type of layer (convolutional, dense, recurrent), the properties of the layer (number of neurons, kernel size, stride size, activation function) and how nodes connect. This is handled by considering a table of possible hyperparameters as the chromosome map for each node and an additional table of global parameters applicable to the entire network (such as learning rate, training algorithm, and data preprocessing) \cite{DBLP:journals/corr/MiikkulainenLMR17}. This makes the algorithm not only define topological information but diverse network configurations more broadly.
        
\begin{figure}[ht]
    \centering
        \includegraphics[width=0.7\textwidth]{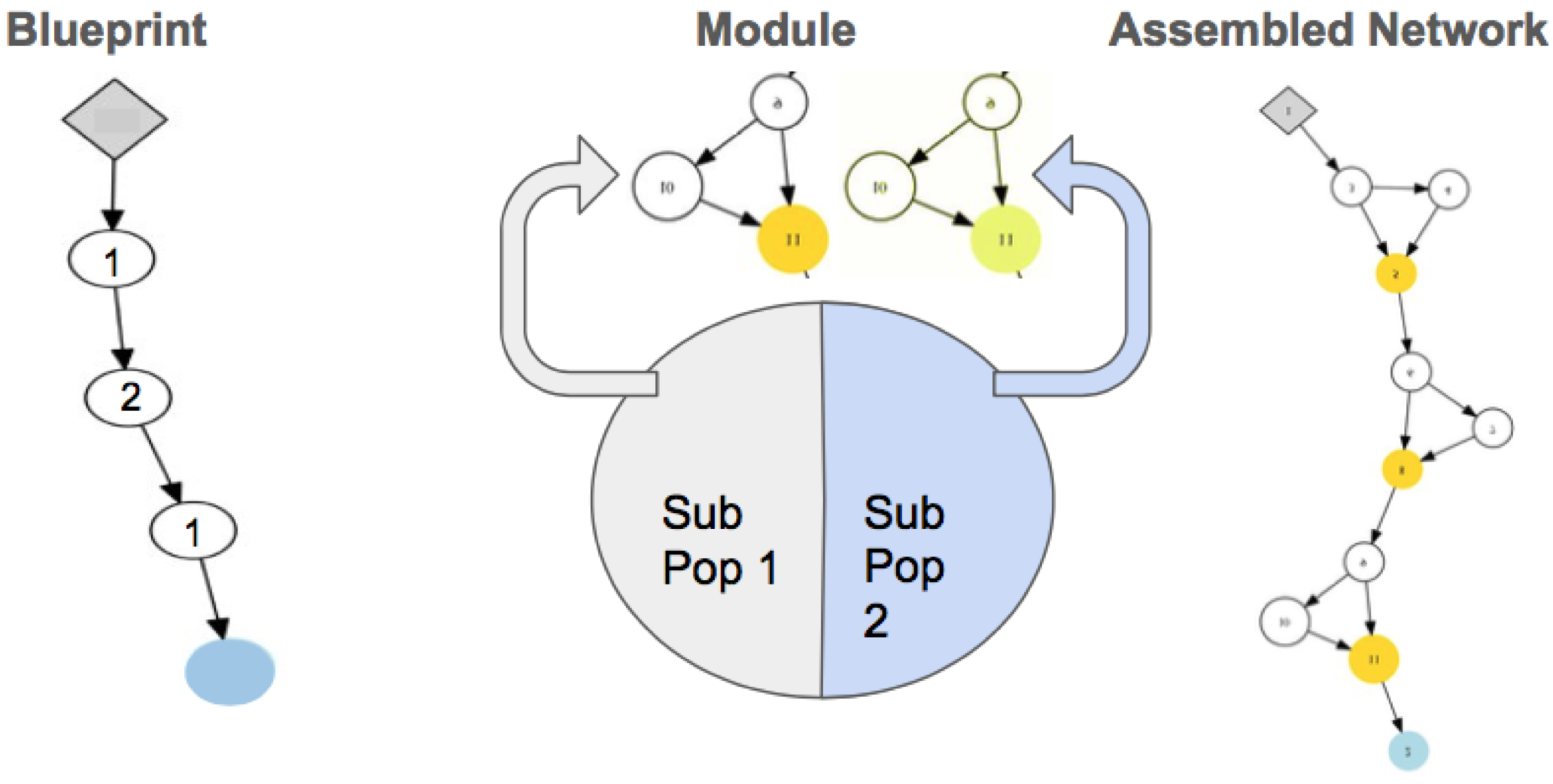}
        \caption{\textbf{CoDeepNEAT network assembling for fitness evaluation.} A visualization of how CoDeepNEAT assembles networks for fitness evaluation. Modules and blueprints are assembled together into a network through replacement of blueprint nodes with corresponding modules. This approach allows evolving repetitive and deep structures seen in many successful recent DNNs. Source: \cite{DBLP:journals/corr/MiikkulainenLMR17}.}        
        \label{fig:codeepneatmoduleblueprint}
\end{figure}

        Investing in the same perspective of focusing on layers instead of single neurons, CoDeepNEAT extends DeepNEAT by dividing the construction of topology into two different levels: module chromosomes and blueprint chromosomes (Figure \ref{fig:codeepneatmoduleblueprint}). Modules are graphs representing a small structure of connected layers. Blueprints are graphs representing a composition of connected nodes that point to module species, which can be assembled into complete networks by joining a sample of the module species pointed by each node. In other words, instead of evolving network species, CoDeepNEAT evolves module species and blueprint species, which are assembled into networks. The algorithm is inspired mainly by Hierarchical SANE \cite{Moriarty1997FormingNN} but is also influenced by the component-evolution approaches called Enforced Sub-populations (ESP) \cite{Gomez:1999:SNC:1624312.1624411} and Cooperative Synapse Neuroevolution (CoSyNE) \cite{gomez:jmlr08}.

Considering these two different chromosome types, CoDeepNEAT requires evolving separate populations for each one of them and scoring them individually. The genetic algorithm behind this is very similar to the one described for DeepNEAT, with the only effective changes being the population management and the assignment of scores by the fitness function. Instead of having one score for each individual and a shared score for the species, the score needs to be assigned to the blueprint and to the modules used in its composition and later shared between their respective species. At the same time, when modules are used in multiple blueprints, all the respective blueprint scores must be considered when assigning a score to a module (averaging them, for instance). Apart from these changes, CoDeepNEAT works very similarly to DeepNEAT while also bringing module evolution as an addition to the standard evaluation process.

The original paper presents results showing that CoDeepNEAT can indeed be implemented and generate high scoring networks for simple datasets such as CIFAR-10 and much more complex problems like image captioning using MSCOCO  \cite{DBLP:journals/corr/ChenFLVGDZ15}. Of course, large datasets require longer training times and more computing resources, which lead CoDeepNEAT to be recently expanded to a platform called Learning Evolutionary AI Framework, or LEAF \cite{2019arXiv190206827L}, taking advantage of cloud computing services to parallelize the algorithm for demanding use cases like pulmonary disease detection on high-resolution chest x-ray images \cite{rajpurkar2017chexnet}.
        
    \subsection{Keras framework} 
    \label{kerassection}
        
        Keras \cite{chollet2015keras} is a popular \footnote{\href{https://towardsdatascience.com/deep-learning-framework-power-scores-2018-23607ddf297a}{https://towardsdatascience.com/deep-learning-framework-power-scores-2018-23607ddf297a}} and high-level neural networks API, written in Python and capable of running on top of TensorFlow \cite{tensorflow2015-whitepaper} and other lower-level frameworks. It was developed with a focus on enabling fast experimentation and allowed for easy and fast prototyping (through user-friendliness, modularity, and extensibility). Keras supports multiple types of neural network components \footnote{\href{https://keras.io/layers/about-keras-layers/}{https://keras.io/layers/about-keras-layers/}}, such as dense layers, convolutional layers, recurrent layers, dropout layers, and supports combinations of them.
        
        The framework automatically manages resources such as CPU and GPU, making efficient use of them. It also has implementations of activation functions \footnote{\href{https://keras.io/activations/}{https://keras.io/activations/}}, optimizers \footnote{\href{https://keras.io/optimizers/}{https://keras.io/optimizers/}}, metric calculations \footnote{\href{https://keras.io/metrics/}{https://keras.io/metrics/}} and procedures needed to manage training sessions with ease.
        
\section{Implementation} \label{chap:implementation}

    The general structure of Algorithm \ref{alg1} is derived directly from the descriptions presented in the original NEAT paper \cite{NEAT}, the CoDeepNEAT paper \cite{DBLP:journals/corr/MiikkulainenLMR17} and its latest implementation, the LEAF platform \cite{2019arXiv190206827L}.
    
    \begin{algorithm}[ht]
    \caption{Genetic algorithm structure for implementation\label{IR}} \label{alg1}
    \SetAlgoLined
    \KwData{hyperparameter tables, global parameter tables}
    \KwResult{evolved candidate solutions}
    \Begin{
    initialize module and blueprint populations considering parameter tables\;
    initialize module and blueprint species\;
     \For{generation in generations}{
        \For{individual in individual population}{
            assemble respective blueprint\;
            generate Keras model\;
            score model\;
            assign score to blueprint and modules\;
        }
        \For{species in module species}{
            calculate shared fitness\;
            apply elitism\;
            reproduce through crossover and mutation considering parameter tables\;
        }
        \For{species in blueprint species}{
            calculate shared fitness\;
            apply elitism\;
            reproduce through crossover and mutation considering parameter tables\;
        }
        speciate modules\;
        speciate blueprints\;
     }
     }
    \end{algorithm}

    Even though the algorithms are described in detail in the original work, a formal pseudo-algorithm is not specified. Thus the procedure described in Algorithm \ref{alg1} is an abstraction of that description. Specific genetic algorithm parameters such as elitism rate, crossover rate, mutation rate, number of allowed species, the minimum and maximum number of individuals per species are not described in depth in the original work. They are implemented as adjustable parameters, as are the tables of possible components of modules (layer types, layer sizes, kernel sizes, strides, activation functions) and hyperparameters (learning rates, optimizers, loss functions).
    
    The general procedures referenced in the algorithm are described in the following subsections, making references to algorithms described in Section \ref{chap:algoschapter}.

    \subsection{Initializing populations} \label{populationinit}
    
        Populations in genetic algorithms are groups of a particular type of individual that will be evolved over generations (Subsection \ref{gasection}). Initializing populations requires a clear description of the involved entities that represent their respective individuals. In the case of the proposed algorithm, these individuals are module entities and blueprint entities, each one being initialized in their respective population (Subsection \ref{codeepneatsection}).
        
        A typical graph design represents module and blueprint entities, only differing in the semantics of their nodes. Even though they represent different levels of abstraction of a single neural network, a standard graph generation procedure generates the graph structures of these entities. As both, they need to follow a shared set of rules designed to project a NN structure correctly. At the same time, they need to respect Keras's limitations when it comes to connecting layers properly.
        
        The graph structures are generated according to the set of rules:
        \begin{itemize}
            \item The base structure of graphs are directed acyclic graphs that map the flow of signals from the input layer to the output layer.
            \item Graphs must follow the limitations established in parameter tables (as showed in Table \ref{table:hyperparamtable}), such as the allowed range of nodes.
            \item Graphs must have exactly one input node and one output node. Both nodes are used to connect graphs to other graphs, despite the internal structure of the graph. This connection takes place in Module to Module connections (in the case of Blueprint graphs) or Layer to Layer connections (in the case of Module graphs). This, in other words, implies the graph can only be connected through its input or its output, not through intermediate nodes.
            \item Nodes in graphs must receive at most two input edges. One input edge directed to an input node means the origin output node can be directly connected to the input node, but more than one input edges connecting to an input node require the inclusion of a merge procedure between them, merging the edges into a single connection, as Layers in Keras can only receive one input signal. For this purpose, Merge layers are implemented in Keras supporting the merging of two input signals each time, meaning that multiple input signals would require constructions of multiple Merge layers. For simplification purposes, this rule guarantees we only have one or two input edges at a node.
            \item Nodes can have multiple output edges. Multiple output signals do not require special treatment.
        \end{itemize}
                
        The graphs are managed in the implementation with the support of NetworkX \cite{SciPyProceedings_11Networkx}, an open-source framework for graph operations in Python, but the rules and graph structure definitions are implemented apart.
        
        Along with structural definitions, graph creation must also handle definitions for the content represented by their nodes and edges. Nodes in these graphs are generated by a routine designed for the creation of node content using custom content creation functions passed as parameters. In the case of modules, which represent assembles of layers, the standard node content creation functions are functions designed to generate new Layers. In the case of blueprints, which represent assembles of modules, the node content creation functions are functions designed to return existing modules from the module population.
        
        For the last part of initializations, individuals are created to represent an instance of a blueprint to be evaluated. Instead of directly evaluating the blueprints, the individuals are instantiated to represent them, because a single blueprint can be trained with different combinations of hyperparameters not associated with the NN structure itself, such as the learning rate, optimizers, loss function or other parameters chosen from a hyperparameter table, explored in Subsection \ref{parametertables}.
        
    \subsection{Parameter tables} \label{parametertables}
    
        Before assembling and training take place, a handful of parameters require management. Mostly addressed as hyperparameters, they relate to decisions made before training starts, like the loss algorithm used, the optimizer for the learning rate, the evaluation metrics to be considered during the training and so on. In this algorithm, additional parameters such as module or blueprint sizes, choices of layer types, configurations of layers, and activation functions can be included in this group.
        
        Table \ref{table:hyperparamtable} exemplifies some of these decisions. The table specifies parameters, the types of decisions required for them, and their range of options, being yet another possible point of optimization in the algorithm. In this specific example, "Module size"  is chosen as a "Random integer,"  ranging from 1 to 3.
        
        \begin{table}[ht]
            \caption{Example hyperparameter table.}
            \begin{center}
                \begin{tabular}{c|c|p{5cm}}
                    \textit{Parameter}  &   \textit{Type}  &   \textit{Options} \\
                    \hline
                    \hline
                    Module size & Random integer & [1, 3] \\
                    Blueprint size & Random integer & [1, 3] \\
                    Component types & Random choice & ["Convolutional", "Dense"] \\
                    Loss functions & Fixed & ["categorical\_crossentropy"] \\
                    Optimizers & Random choice & ["Adam", "RMSprop"] \\
                    Evaluation metrics & Fixed & ["Accuracy"]\\
                    \hline
                \end{tabular}
            \end{center}
            \label{table:hyperparamtable}
        \end{table}
        
        Similarly, specific component tables can be established to define the configuration of layers during the module constructions. Table \ref{table:convtable} exemplifies the parameters considered during the instantiation of a convolutional layer. For instance, the table specifies that any convolutional layer will need to range their "Filters"  as a "Random integer"  between 32 and 64 while choosing "Kernel sizes"  among 3, 5, and 7.
        
       Another possible point of optimization in the algorithm, these tables could be evolved in their populations during generations, similarly to modules and blueprints. This would allow the improvement of the usage of different hyperparameters in the training of different individuals, evaluating the table setups, and evolving them over time. Though the current implementation only uses fixed tables as the ones represented in \ref{table:hyperparamtable} and \ref{table:convtable}, a similar hyperparameter optimization procedure is implemented in LEAF \cite{2019arXiv190206827L}, adding an additional dimension to the evolution process.

        \begin{table}[ht]
            \caption{Example component parameter table for convolutional layers.}
            \begin{center}
                \begin{tabular}{c|c|p{5cm}}
                    \textit{Parameter}  &   \textit{Type}  &   \textit{Options} \\
                    \hline
                    \hline
                    Filters & Random integer & [32, 64]\\
                    Kernel size & Random choice & [3, 5, 7]\\
                    Stride & Random choice & [2, 3]\\
                    Activation function & Random choice & ["relu", "tanh"]\\
                    Dropout & Random float & [0, 0.7]\\
                    \hline
                \end{tabular}
            \end{center}
            \label{table:convtable}
        \end{table}

    \subsection{Initializing and managing species}

        Speciation plays an essential role in NEAT and its variants, ensuring the diversity inside populations over generations, as described in Subsection \ref{neatsection}. The species must be initialized along with the populations for relevant procedures like elitism, crossover, and mutation take place.
        
        The method used to approximate module and blueprint similarity to generate species is K-means. The original paper for CoDeepNEAT comments their usage of the same speciation schemes used for NEAT but does not specify in detail how these schemes translate when dealing with the different representations of individuals used in CoDeepNEAT. This specific part was abstracted and implemented in this work using K-Means, whose functionality is described in Subsection \ref{clusteringsection}.
        
        K-Means is used to cluster module and blueprint graphs based on three main structural pieces of information: 
        \begin{itemize}
            \item the size of the network, such as the sum of the number of filters in convolutional layers of neurons in fully-connected layers, or simply the number of neuron connections;
            \item count of nodes, representing the number of layers or modules in the graph;
            \item count of edges, representing the number of connections between nodes in the graph.
        \end{itemize}
         
        This choice of clustering features can be easily changed in the implementation, as it is merely a parameter for K-means. This specific set of features evaluates only quantitative information, but qualitative information such as training scores (which would require training before an evaluation) or other types of scores could be used. The clusterization generates an automatic (or custom) amount of clusters, which are used as species.
        The k-means implementation used is from the open-source framework Scikit-Learn \cite{scikit-learn}.

        After species initialization, the nearest centroid method explained in Subsection \ref{clusteringsection} is used to assign new members to an existing species, allowing species to grow and change over generations. Centroids are calculated based on the features of the current species members every time new members need to be assigned to a species. New members are then assigned to the closest centroid. This way, members that already have an assigned species (e.g., members kept by elitism or new members that were assigned a species by any other method) still belong to their original species, but entirely new members are assigned to the adequate species according to the species' current demographic. An accuracy threshold can be specified, so new species are generated in case new members do not fit the existing centroid with satisfactory proximity.
        The nearest centroids implementation used is from the open-source framework Scikit-Learn \cite{scikit-learn}.
        
    \subsection{Neural network assembling}
    
        Transitioning the graphs to Keras model representations is required to take profit from the training procedures available in the Keras framework. After modules and blueprints are created, they need to be assembled into a unique graph, which is subsequently processed, node by node, creating the respective Keras layers and connecting them one by one following a topological sorting \footnote{\href{https://networkx.github.io/documentation/stable/reference/algorithms/dag.html}{https://networkx.github.io/documentation/stable/reference/algorithms/dag.html}}.
        
\begin{figure}[ht!]
            \centering
            \begin{minipage}[b]{.45\textwidth}
                \centering
                \begin{subfigure}[b]{\linewidth}
                    \centering
                    \includegraphics[width=3.4cm]{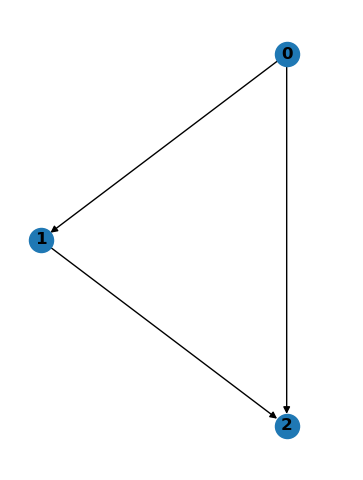}
                    \caption{Module graph structuring the connections of 3 different layers.}
                    \label{fig:example_network/1_2_intermed_module}
                \end{subfigure}
                \begin{subfigure}[b]{\linewidth}
                    \centering
                    \includegraphics[width=3.4cm]{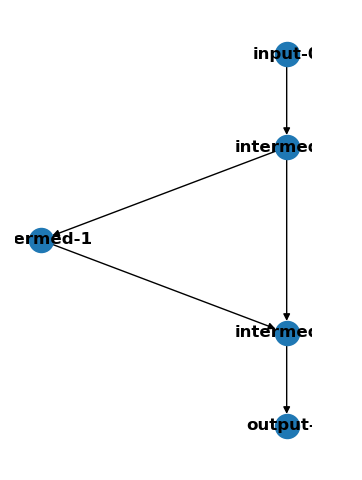}
                    \caption{Blueprint graph structuring the connections of 3 instances of module (a), plus additional input and output layers.}
                    \label{fig:example_network/1_module_level_graph}
                \end{subfigure}
                \begin{subfigure}[b]{\linewidth}
                    \centering
                    \includegraphics[width=3.4cm]{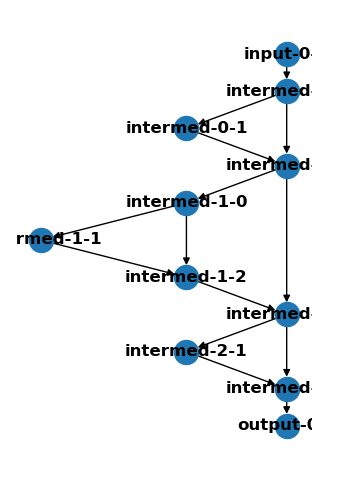}
                    \caption{Graph assembling the module (a) and blueprint (b) information into one structural representation.}                    
                    \label{fig:example_network/2_component_level_graph}
                \end{subfigure}
            \end{minipage}
            \begin{minipage}[b]{.5\textwidth}
                \begin{subfigure}[b]{\linewidth}
                    \includegraphics[width=6.5cm]{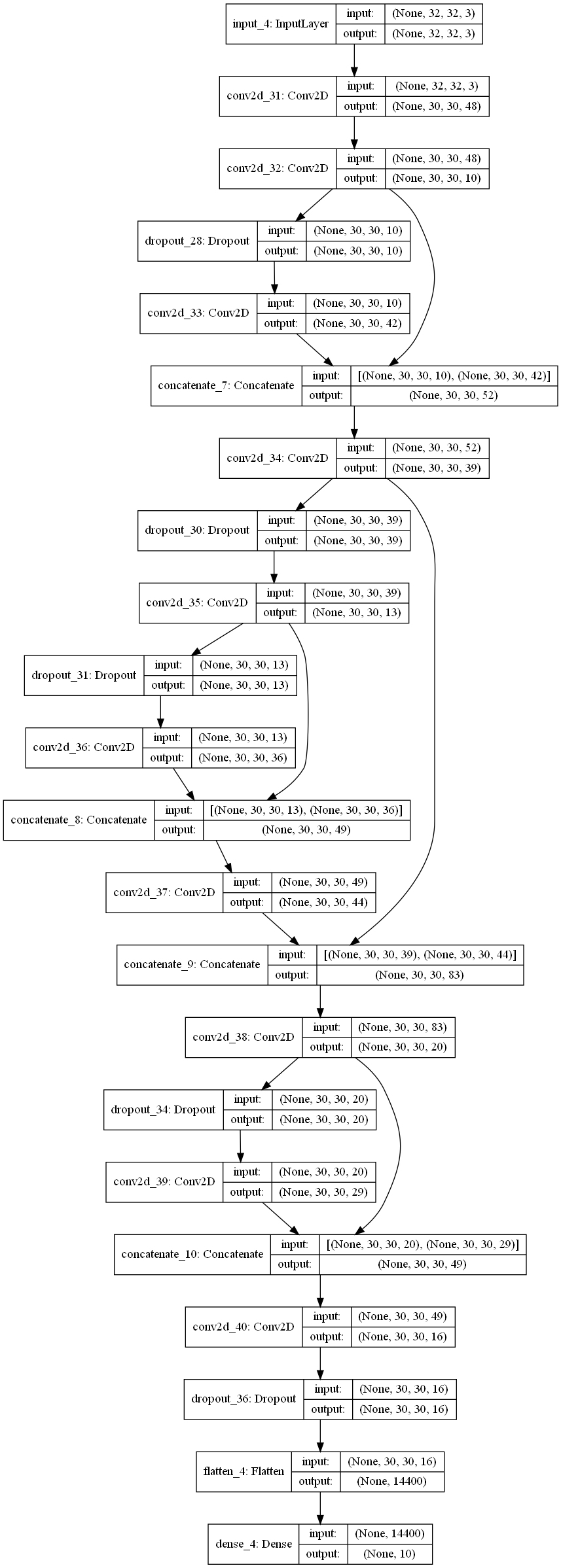}
                    \caption{Final Keras model representation generated from the assembled graph (c).} 
                    \label{fig:example_network/3_layer_level_graph}
                \end{subfigure}
            \end{minipage}
            \caption{\textbf{Different views on the assembling of a Network.}}
            \label{fig:example_network}
        \end{figure}

        The transition scheme implemented handles the necessary interactions between layers, such as including Merge layers between two inputs directed to one layer, or adding Flatten layers to adjust the connections between Convolutional and Dense layers \footnote{\href{https://keras.io/layers/core/}{https://keras.io/layers/core/}}. After the model is completely connected and a set of hyperparameters is set, it is trained using the standard Keras training functions and a specified dataset. An assembled graph containing joined blueprint and module information can be seen in Figure \ref{fig:example_network/2_component_level_graph} and its respective network can be seen in Figure \ref{fig:example_network/3_layer_level_graph}.
        
        Figure \ref{fig:example_network/3_layer_level_graph} shows an example assembled Keras network generated by the algorithm. This network is based on the assembled graph shown in Figure \ref{fig:example_network/2_component_level_graph}, which is structured by the network's blueprint shown in Figure \ref{fig:example_network/1_module_level_graph}. Figure \ref{fig:example_network/1_2_intermed_module} shows the graph structure for the common module used by the three intermediate nodes in the blueprint graph in Figure \ref{fig:example_network/1_module_level_graph}.

        The resulting scores from the Keras scoring procedures are then extracted from the trained model and assigned to the individual and its respective blueprint, which propagates them to the underlying modules involved. This score is used in the evaluation procedures to decide which entities survive elitism and which entities are candidates for reproduction in crossover or mutation schemes.

    \subsection{Elitism, crossover, and mutations}
    
        The algorithm handles population management procedures such as elitism, crossover, and mutations after the current populations are evaluated by training and scores. Currently, the proportional amount of subjects to these procedures needs to be defined by the user but could, alternatively, be explored and evolved in the hyperparameter tables.
        
        The implemented elitism mechanism follows the standard definition of the concept, preserving a certain percentage of individuals in populations through generations, ensuring the survival of the best solutions.
        
\begin{figure}[ht!]
            \vspace{.05\textwidth}
            \centering
            \begin{minipage}[t]{.45\textwidth}
                \begin{subfigure}[t]{\linewidth}
                    \centering
                    \includegraphics[width=4.4cm]{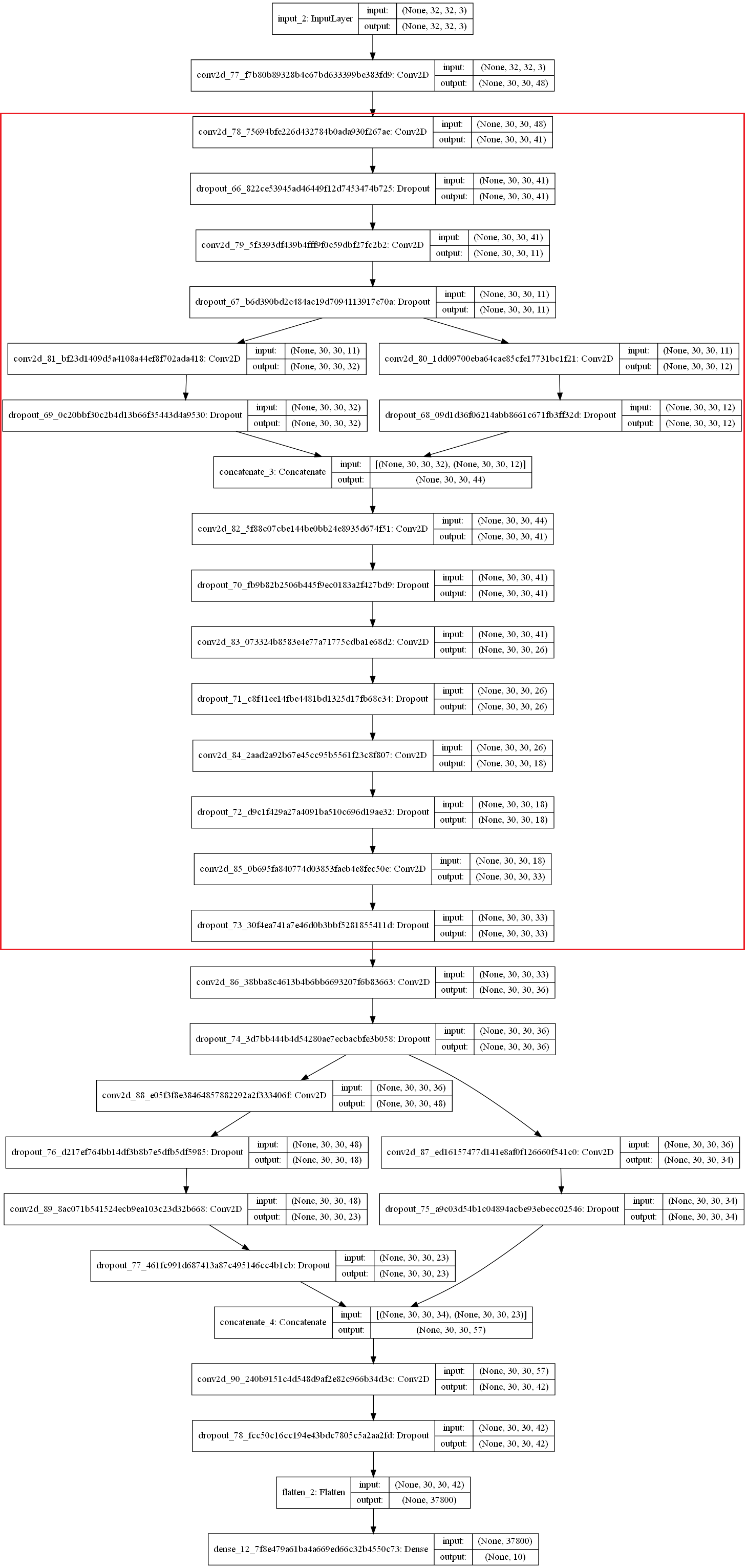}
                      \caption{Parent 1 network.}                    
                    \label{fig:crossover/parent1}
                \end{subfigure}
            \end{minipage}
            \begin{minipage}[t]{.45\textwidth}
                \begin{subfigure}[t]{\linewidth}
                    \centering
                    \includegraphics[width=4.4cm]{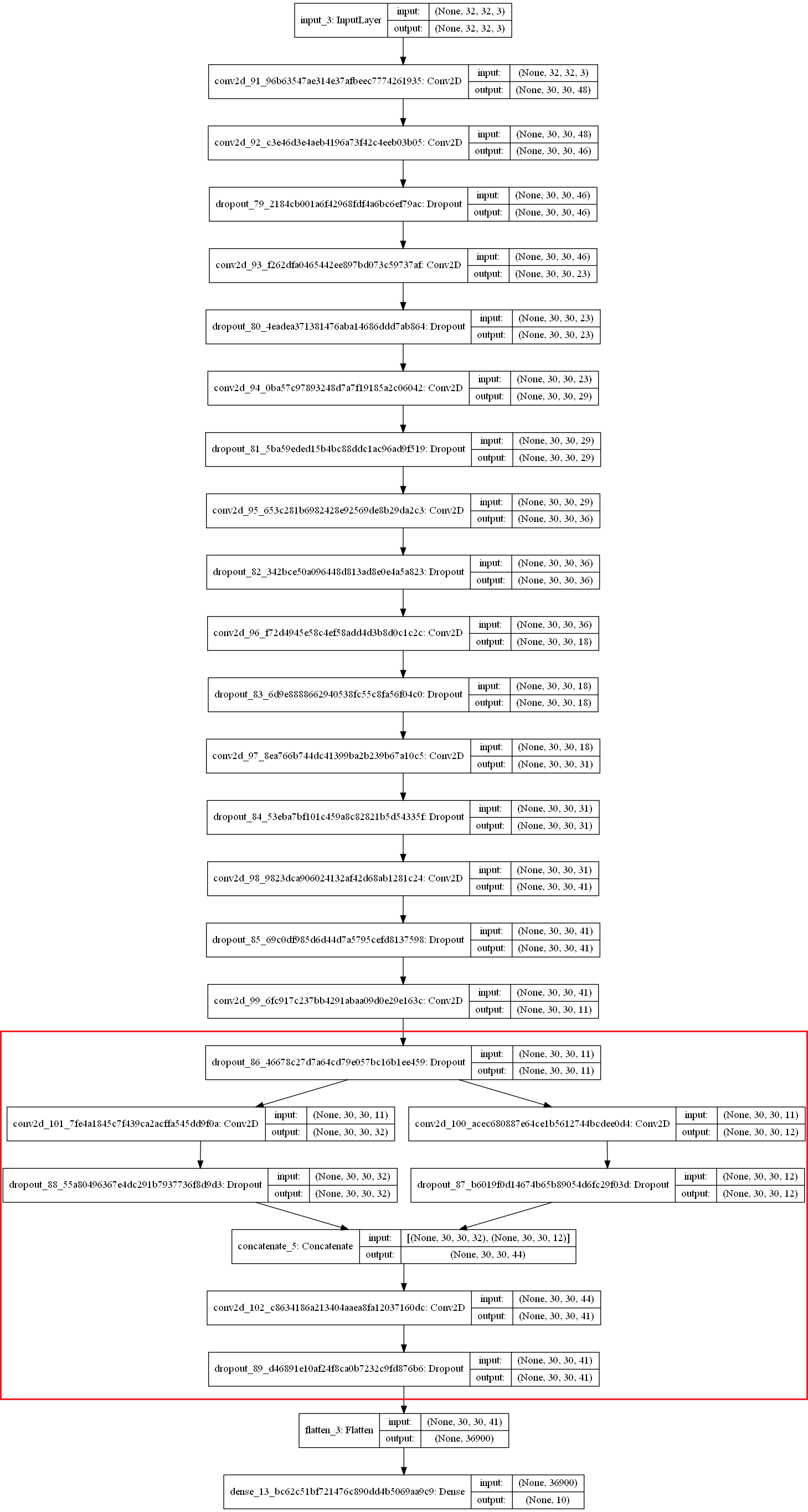}
                    \caption{Parent 2 network.}                    
                    \label{fig:crossover/parent2}
                \end{subfigure}
            \end{minipage}
            \\
            \begin{minipage}[t]{.45\textwidth}
                \begin{subfigure}[t]{\linewidth}
                    \centering
                    \includegraphics[width=4.4cm]{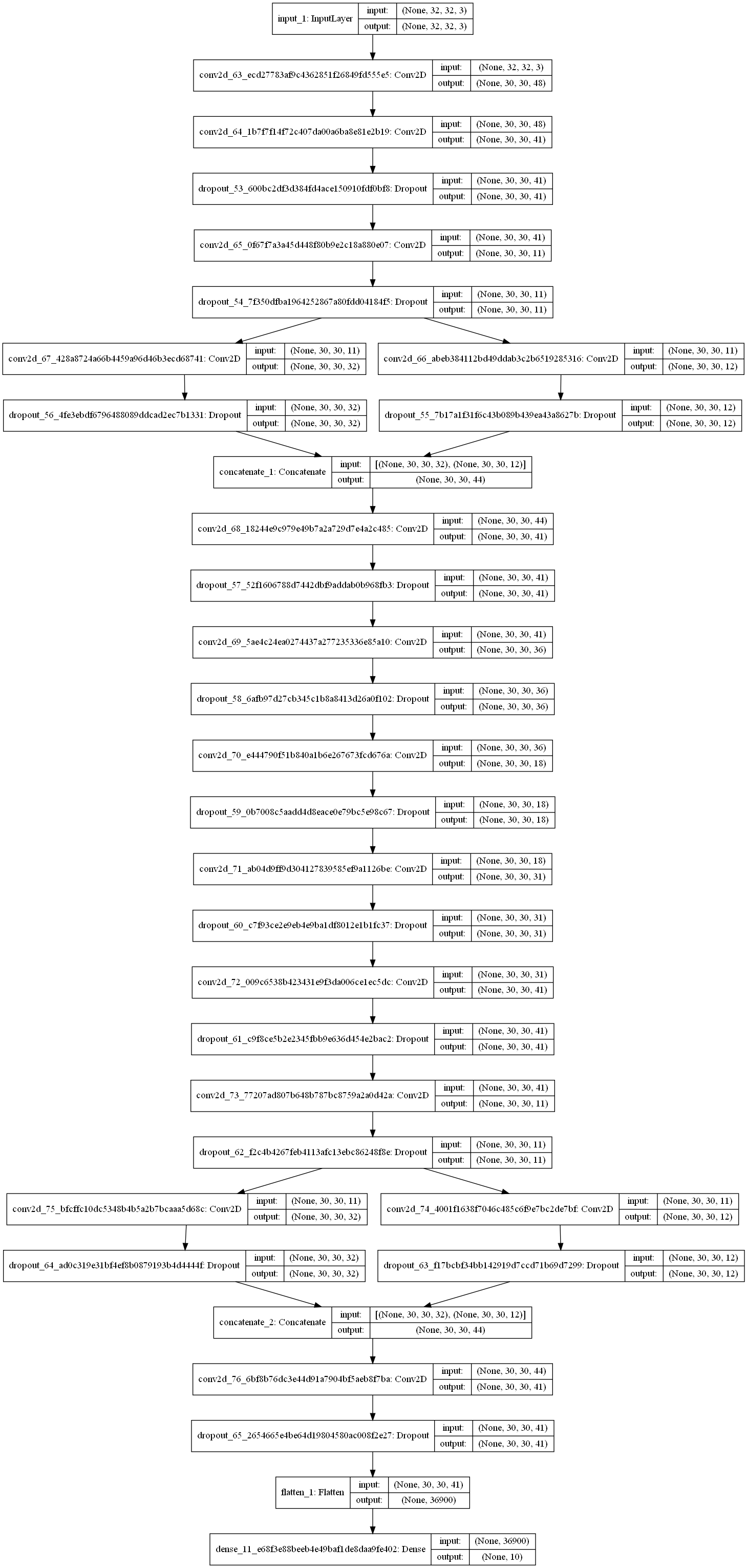}
                    \caption{Child network.}                    
                    \label{fig:crossover/child}
                \end{subfigure}
            \end{minipage}
            \caption{\textbf{A crossover example.} Genetic information from Parent 1 (a) and Parent 2 (b), highlighted in red, are combined to generate a new network (c). The figures depict the network representation of the three blueprints involved in the crossover process.}
            \label{fig:crossover/network}
        \end{figure}

        Crossover is implemented following a uniform crossover technique \cite{Holland:1975}, switching node contents of two graphs with a fixed chance. The effects are different in blueprints and modules, representing whole layer connection switches in the first case, and simply layer definition switches in the latter. A visual representation of the crossover operation effects can be seen in Figure \ref{fig:crossover/network}, where a complete network is generated from two-parent networks. The crossover handles cases where layers or modules are not compatible to be switched by only exchanging regions with common origins, as in NEAT.

        Mutations are implemented similarly to the original proposal of NEAT, representing edge and node alterations such as a node or edge removal, creation, or reconnection. The main differences when comparing to what is proposed by CoDeepNEAT are that while NEAT represents the node content as activation functions and edge contents as weights, CoDeepNEAT's representations of node content are much more complicated (NN structures!), implying that more complex interactions need to be considered. For this reason, the mutations must also follow the same set of rules as specified for graphs in the population's initializations subsubsection (\ref{populationinit}).
        
\begin{figure}[ht!]
            \centering
            \begin{minipage}[t]{.49\textwidth}
                \begin{subfigure}[b]{\linewidth}
                    \centering
                    \includegraphics[width=3.5cm]{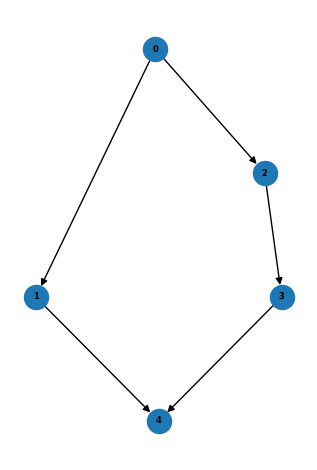}
                    \caption{Original graph.}
                    \label{fig:crossover/original}
                \end{subfigure}
                \begin{subfigure}[b]{\linewidth}
                    \centering
                    \includegraphics[width=3.5cm]{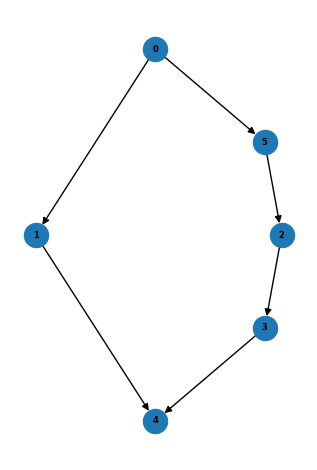}
                    \caption{Mutation by node addition.}                    
                    \label{fig:crossover/mutation_by_node_addition}
                \end{subfigure}
            \end{minipage}
            \begin{minipage}[t]{.49\textwidth}
                \begin{subfigure}[b]{\linewidth}
                    \centering
                    \includegraphics[width=3.5cm]{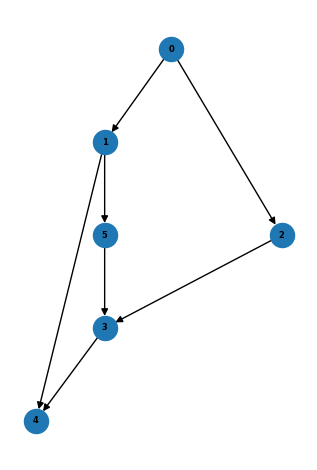}
                    \caption{Mutation by node addition.}                    
                    \label{fig:crossover/mutation_by_node_addition_2}
                \end{subfigure}
                \begin{subfigure}[b]{\linewidth}
                    \centering
                    \includegraphics[width=3.5cm]{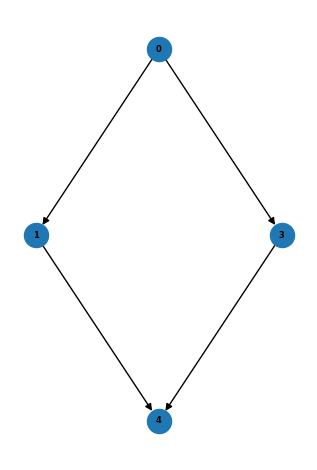}
                    \caption{Mutation by node removal.}                    
                    \label{fig:crossover/mutation_by_removal}
                \end{subfigure}
            \end{minipage}
            \caption{\textbf{Mutation examples.} Figure (a) shows the original graph before any mutations take place. Figure (b) shows the mutation of the original graph by inserting a node between an existing connection. Figure (c) shows the Mutation of the original graph adding a node and preserving existing edges. Figure (d) shows the mutation of the original graph by removing an existing node.}
            \label{fig:example_mutations}
        \end{figure}

\begin{figure}[ht!]
            \centering
            \begin{minipage}[]{.29\textwidth}
                \begin{subfigure}[t]{\linewidth}
                    \centering
                    \includegraphics[width=0.8\linewidth]{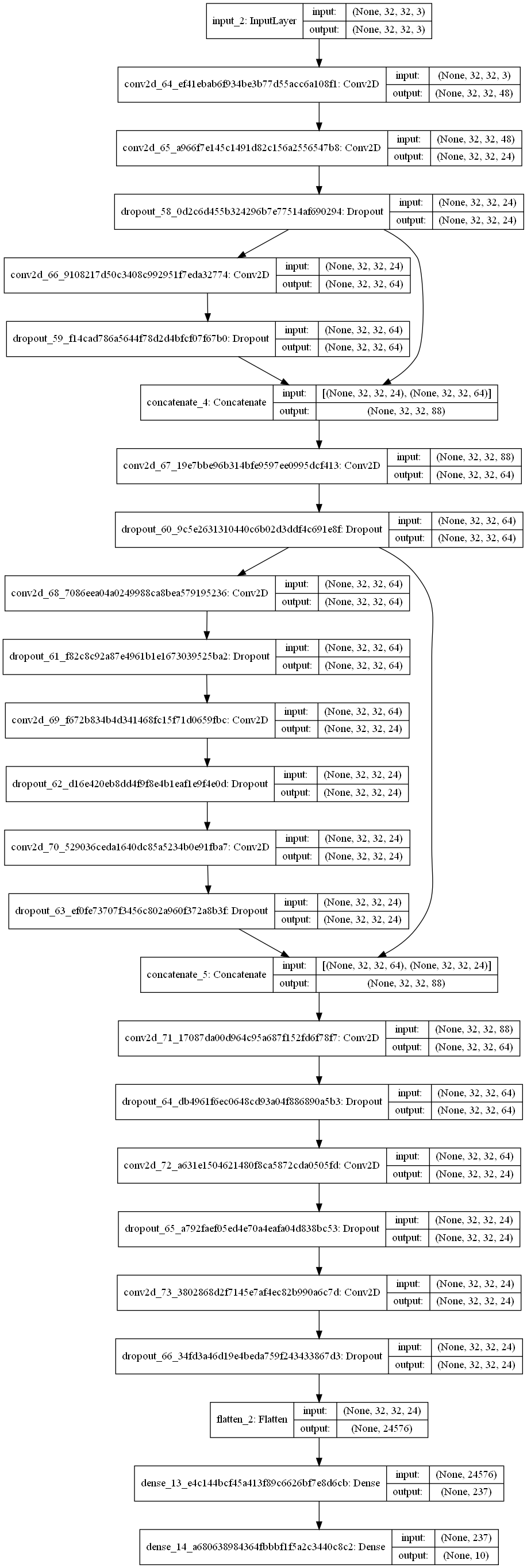}
                    \caption{Original network.}
                    \label{fig:crossover/mutation_content/model}
                \end{subfigure}
            \end{minipage}
            \begin{minipage}[]{.29\textwidth}
                \begin{subfigure}[t]{\linewidth}
                    \centering
                    \includegraphics[width=.4\linewidth]{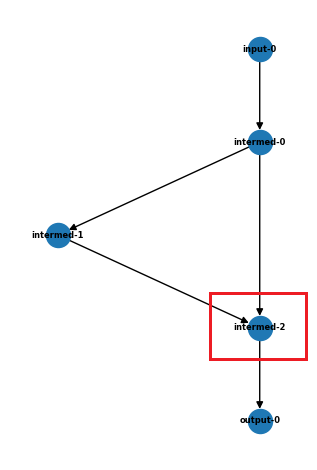}
                    \caption{Original blueprint.}                    
                    \label{fig:crossover/mutation_content/blueprint}
                \end{subfigure}
                \begin{subfigure}[t]{\linewidth}
                    \centering
                    \includegraphics[width=.4\linewidth]{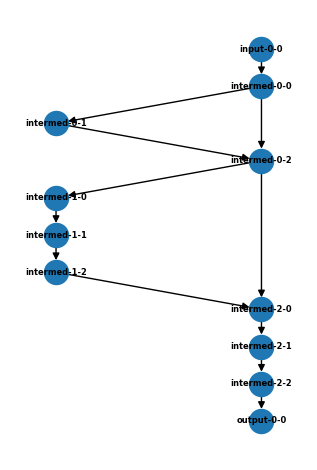}
                    \caption{Original assembled graph.}                    
                    \label{fig:crossover/mutation_content/original_assembled}
                \end{subfigure}
                \begin{subfigure}[t]{\linewidth}
                    \centering
                    \includegraphics[width=.4\linewidth]{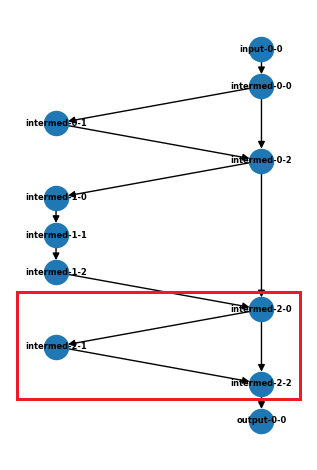}
                    \caption{Assembled graph after mutating the blueprint node.}     
                    \label{fig:crossover/mutation_content/mutated_assembled}
                \end{subfigure}
            \end{minipage}
            \begin{minipage}[]{.29\textwidth}
                \begin{subfigure}[t]{\linewidth}
                    \centering
                    \includegraphics[width=0.8\linewidth]{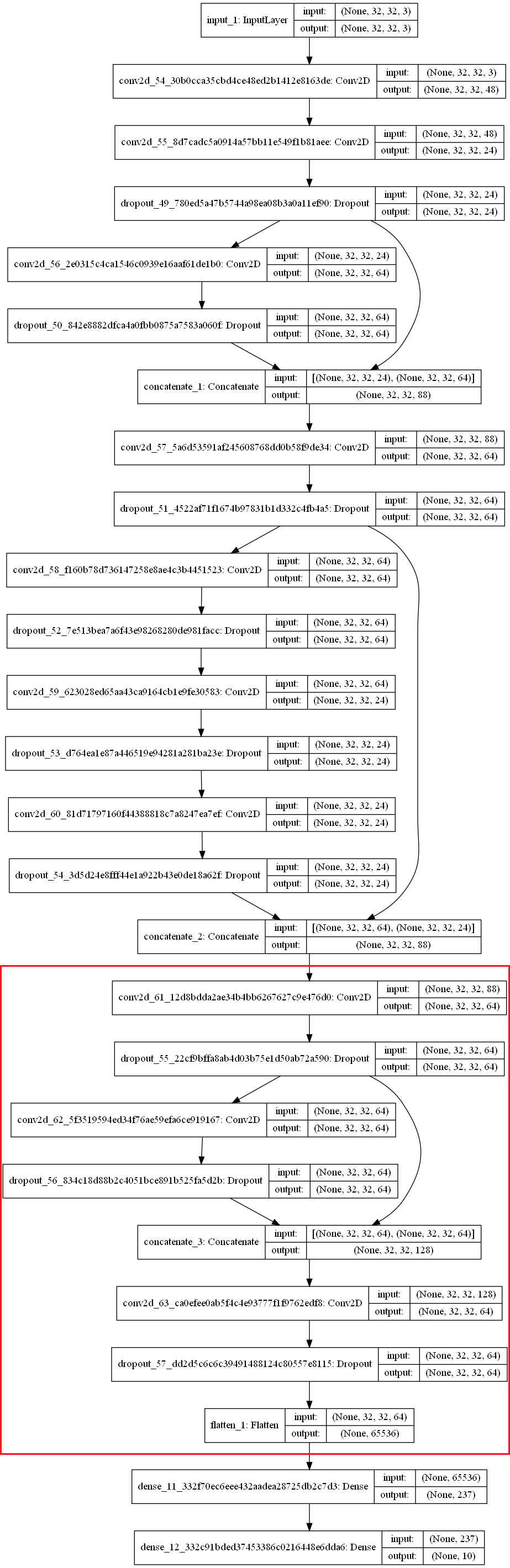}
                    \caption{Mutated network.}                    
                    \label{fig:crossover/mutation_content/mutated_model}
                \end{subfigure}
            \end{minipage}
            \caption{\textbf{Node content mutation effects on a network.} (a) Original assembled network. (b) The original blueprint used in the network and the indicated node (a module) to be replaced, highlighted in red. (c) The assembled graph of the used blueprint before mutation switches the indicated module. (d) Assembled graph of the blueprint after mutation switches the indicated module. (d) Resulting network from mutation, with the affected part highlighted in red.}
            \label{fig:example_mutation_content}
        \end{figure}

        Mutations take place in the graph representations of modules and blueprints as structural changes, as represented in Figures \ref{fig:example_mutations} and \ref{fig:example_mutation_content}. Mutation operators are implemented as: 
        
        \begin{itemize}
            \item Node additions, creating nodes and connecting them to other nodes either by inserting them between two nodes that already have an existing mutual connection (Figure \ref{fig:crossover/mutation_by_node_addition}), or by connecting them to any pair of nodes that support new connections (Figure \ref{fig:crossover/mutation_by_node_addition_2}).
            \item Node removals, creating a new connection between the direct neighbors of the former node (Figure \ref{fig:crossover/mutation_by_removal}).
            \item Edge removals or additions.
            \item Node replacement, changing current node content, such as replacing modules in blueprint graphs (as in Figure \ref{fig:example_mutation_content}) or layers in module graphs.
        \end{itemize}
        
        The results from these changes can be seen mostly in the final representations of the models when the assembling process is finished. In Figure \ref{fig:example_mutation_content}, a node content change in the original blueprint of the network results in significant changes to its structure after the assembling process.
        
\section{Experiments and Results}\label{chap:experiments}

    Experimentation on the algorithm followed consecutive executions in two different image datasets. This Section describes datasets, experiments, and results, as well as discusses the results and practical usability of the algorithm.
    
    \subsection{Datasets}
    
        The chosen datasets for experiments using this implementation were MNIST \cite{MNIST} and CIFAR-10 \cite{CIFAR-10}.
        Both datasets are simple image datasets containing ten different classes of images.
        They are frequently used in benchmarks or experiments using convolutional or fully-connected networks and have been used before in the task of topology selection in other approaches \cite{DBLP:journals/jece/MattioliCCNL19}.
        
        The parameter tables used in both experiments can be seen in Table \ref{table:cifar_experiment_hyperp}. The number of modules and layers used for blueprint and module constructions, respectively, are specified to be in the range between 1 and 3. The intention is to build minimal structures at first and progressively grow these structures as mutations and crossovers take place as generations pass. Convolutional layers are specified to be used in the intermediate layers, while dense layers are used in the output layers. Including dense layers in the last layer before outputs is a common practice in successful convolutional networks \cite{DBLP:journals/corr/SimonyanZ14a}. Tables \ref{table:cifar_experiment_conv} and \ref{table:cifar_experiment_dense} specify the possible configurations of these two layer types.
        
        \begin{table}[ht]
            \caption{Experiment hyperparameter table.}
            \begin{center}
                \begin{tabular}{c|c|p{5cm}}
                    \textit{Parameter}  &   \textit{Type}  &   \textit{Options} \\
                    \hline
                    \hline
                    Module size & Random integer & [1, 3] \\
                    Blueprint size & Random integer & [1, 3] \\
                    Intermediate component types & Fixed & ["Convolutional"] \\
                    Output layer component types & Fixed & ["Dense"] \\
                    Loss functions & Fixed & ["categorical\_crossentropy"] \\
                    Optimizers & Fixed & ["Adam"] \\
                    Evaluation metrics & Fixed & ["Accuracy"]\\
                    \hline
                \end{tabular}
            \end{center}
            \label{table:cifar_experiment_hyperp}
        \end{table}
        \begin{table}[ht]
            \caption{Experiment parameter table for Convolutional layers.}
            \begin{center}
                \begin{tabular}{c|c|p{5cm}}
                    \textit{Parameter}  &   \textit{Type}  &   \textit{Options} \\
                    \hline
                    \hline
                    Filters & Random integer & [16, 48]\\
                    Kernel size & Random choice & [1, 3, 5]\\
                    Stride & Fixed & [1]\\
                    Activation function & Fixed & ["relu"]\\
                    Dropout & Random float & [0, 0.5]\\
                    \hline
                \end{tabular}
            \end{center}
            \label{table:cifar_experiment_conv}
        \end{table}
        \begin{table}[ht]
            \caption{Experiment parameter table for Dense layers.}
            \begin{center}
                \begin{tabular}{c|c|p{5cm}}
                    \textit{Parameter}  &   \textit{Type}  &   \textit{Options} \\
                    \hline
                    \hline
                    Units & Random integer & [32, 256]\\
                    Activation function & Random choice & ["relu"]\\
                    \hline
                \end{tabular}
            \end{center}
            \label{table:cifar_experiment_dense}
        \end{table}
        
    \subsection{MNIST experiment}
    
        Initial experimentation took place using MNIST, a fast-converging and widely used dataset in handwritten digit recognition tasks and overall convolutional network experiments \cite{WittenFrankHall11}. MNIST is composed of 60000 28x28 pixel grayscale images of handwritten numerical digits divided into ten classes \cite{MNIST}. The images are simple and placed in neutral backgrounds that simplify predictions.
        
        \begin{figure}[ht]
            \centering
            \includegraphics[width=0.6\textwidth]{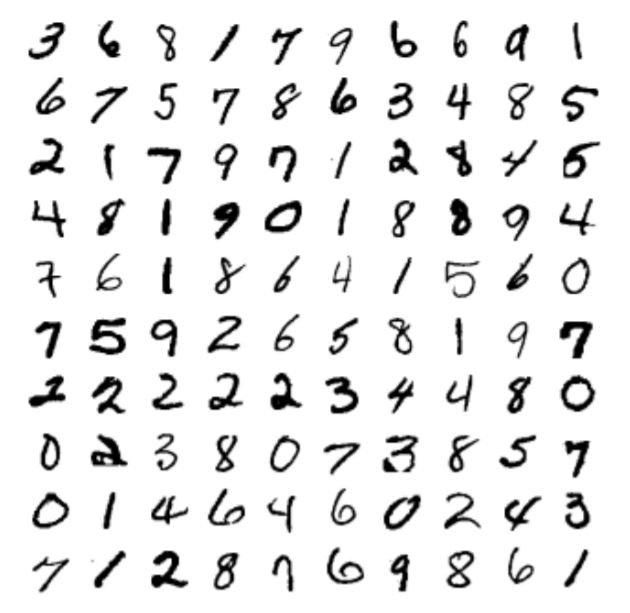}
            \caption{\textbf{Samples from the MNIST dataset, a handwritten numerical digit dataset.} Source: \cite{MNIST}.}
            \label{fig:mnist_dataset}
        \end{figure}
        
        Experimentation with MNIST was done using 40 generations, populations of 10 individuals, 10 blueprints, and 30 modules, as well as a starting number of species set to 3. For each population, a global set of configurations was used to define elitism, crossover, and mutation rates. The elitism rate is set to 20\%, preserving the best-scoring solutions every generation. The crossover rate is set to 30\%, replacing the same proportion of populations' individuals with offspring from good scoring parents. The remaining 50\% of the population is subject to mutation operations, generating random changes to existing solutions.
        
        Training using MNIST usually divides the dataset into three parts: a training dataset, composed of 42500 images; a validation dataset, composed of 7500 images; and a test dataset, composed of 10000 images. For this type of experiment, training the network to its full length is a hardware and time-consuming task. Topology selection methods commonly reduce the sizes of these datasets to smaller proportions to achieve faster results and discard bad solutions in early generations, avoiding the waste of resources in lengthy training procedures, as in \cite{DBLP:journals/jece/MattioliCCNL19}. For this reason, the training sessions over generations used random samples of 10000 images from the original 60000 divided into 8000 training samples and 2000 validation samples.
        
\begin{figure}[!ht]
            \centering
            \includegraphics[width=15cm]{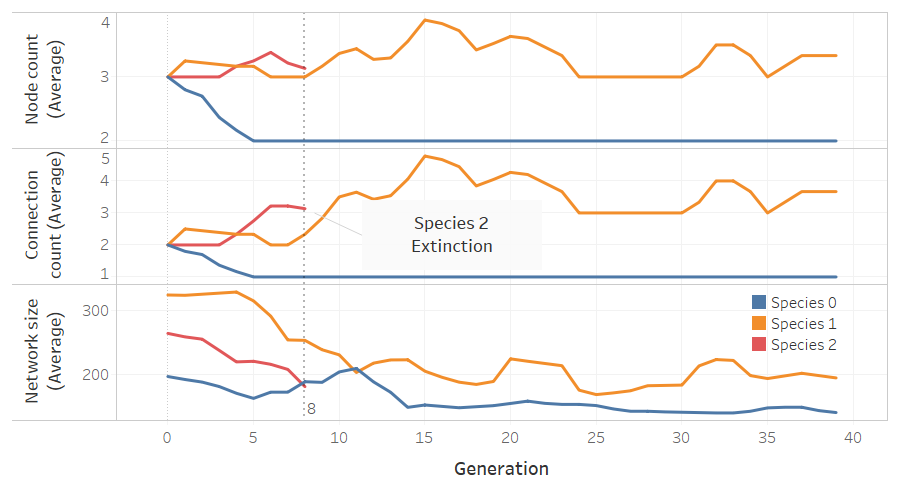}      \caption{\textbf{Progress of the features of the three species of networks generated for MNIST over generations.} The representation shows the average of each feature for each species. The different line colors represent different species.}
            \label{fig:mnist_species_features}
        \end{figure}

        As mutation and crossover operations take place along generations, the features of the networks are expected to change and adapt to reach better accuracy and loss scores during early training. At the same time, elitism ensures these operations do not change actual good results. Figure \ref{fig:mnist_species_features} depicts the changes in the counts of nodes, connections, and the overall network size of the blueprint population as the generations pass.
        
        In the experiment history shown in Figure \ref{fig:mnist_species_features}, the networks tend to decrease in size even when increasing the number of nodes or connections (as in Species 1). This means that the networks are using fewer filters or neurons in their layers, which is expected behavior due to MNIST being a straightforward dataset that does not require complex structures to achieve high loss and accuracy metrics \cite{MNIST}. The reduced dataset sizes and training epochs used in topology selection also tends to favor fast-converging networks, as seen in the experiments of \cite{DBLP:journals/corr/MiikkulainenLMR17}. Also, Figure \ref{fig:mnist_species_features} shows how the evolution of features results in the populations taking certain paths, leading some species (in this case, Species 2) to eventually cease to have representatives, even when not directly interacting with other species through crossovers.
        
\begin{figure}[ht!]
            \centering
            \includegraphics[width=15cm]{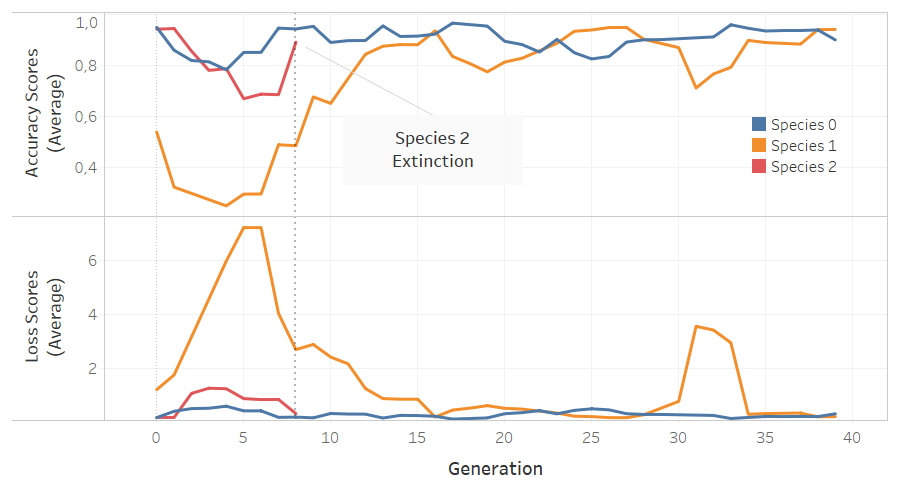}
            \caption{\textbf{Progress of the average accuracy and loss scores of the species over generations for MNIST dataset.} The different colors represent different species.}
            \label{fig:mnist_species_scores}
        \end{figure}

        Changes and adaptations to the network's features result in changes to the species scores (Figure \ref{fig:mnist_species_scores}), reducing the loss metrics and increasing the accuracy metrics.
        
\begin{figure}[ht]
            \begin{minipage}[t]{.49\textwidth}
                \begin{subfigure}[b]{\linewidth}
                    \includegraphics[width=7.5cm]{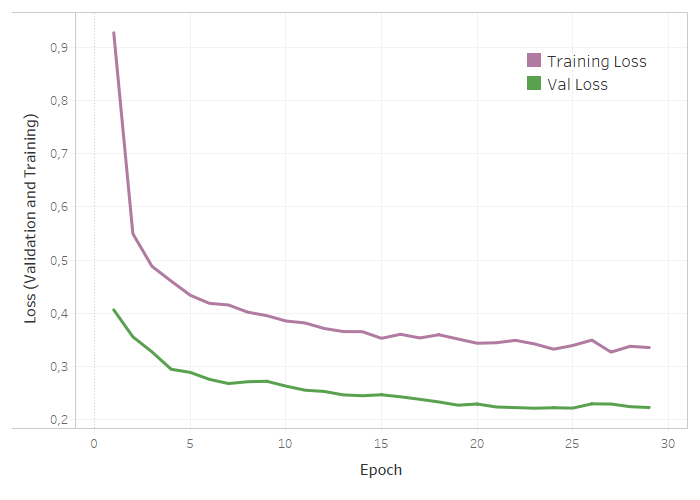}
                    \caption{Loss.}
                    \label{fig:mnist_loss}
                \end{subfigure}
            \end{minipage}
            \begin{minipage}[t]{.49\textwidth}
                \begin{subfigure}[b]{\linewidth}
                      \includegraphics[width=7.5cm]{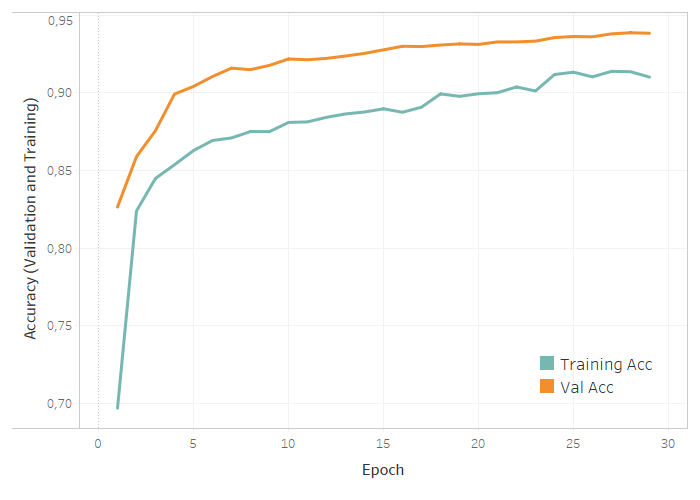}
                      \caption{Accuracy.}  
                    \label{fig:mnist_acc}
                \end{subfigure}
            \end{minipage}
            \caption{\textbf{Training and validation metrics for the best network generated for MNIST after 40 generations.}}
            \label{fig:mnist_metrics}
        \end{figure}

        After the 40 generations, a network was selected by the highest accuracy and loss scores. The chosen network was then trained using the complete MNIST training dataset for 30 epochs to validate whether it would generate acceptable results or not. The training metrics are shown in Figure \ref{fig:mnist_metrics} and demonstrate that even in the early epochs, the resulting model achieves more than 90\% validation accuracy. The accuracy using the test dataset achieved a peak of 92\% accuracy at epoch 30. Of course, the MNIST dataset is supposed to be easy to predict and achieve very high accuracy metrics (98.5\% \footnote{\href{https://towardsdatascience.com/image-classification-in-10-minutes-with-mnist-dataset-54c35b77a38d}{https://towardsdatascience.com/image-classification-in-10-minutes-with-mnist-dataset-54c35b77a38d}}, for instance). This result shows that the algorithm was able to achieve an acceptable result with a few generations, even though the initial network size could be smaller.
        
    \subsection{CIFAR-10 experiment}
    
\begin{figure}[ht!]
            \centering
            \includegraphics[width=0.9\textwidth]{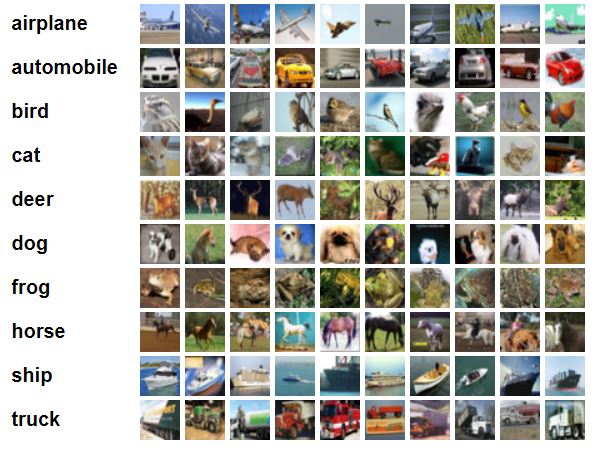}
            \caption{\textbf{Samples from the CIFAR-10 dataset \cite{CIFAR-10}, a collection of different classes of images in small scale.}} 
            \label{fig:cifar_dataset}
        \end{figure}

        Experimentation continued using CIFAR-10, a slightly more complex dataset than MNIST. CIFAR-10 is composed of 60000 32x32 colored images of different objects divided in 10 classes (Figure \ref{fig:cifar_dataset}). Even though CIFAR-10 is similar to MNIST in sample quantity and size, its images represent much more complex and diverse object structures for each class when comparing to MNIST. Training is more exhausting, as well as the required model structure for better results is usually bigger \cite{CIFAR-10}.

        For CoDeepNEAT's original CIFAR-10 experiment, the authors describe the execution of 72 generations using populations of 25 blueprints and 45 modules to generate 100 individuals (CNNs) per generation. The evaluation of these individuals is done through the test scores of their respective CNNs after eight training epochs using 50000 images divided into a training set of 42500 samples and a validation set of 7500 samples. Since training convolutional neural networks takes a long time, the reduced training epochs are necessary to achieve approximations of adequate topologies in a viable time. Still, the processing required to train all the individuals in every generation for multiple generations is considerable.
        
        After the evolution process of CoDeepNEAT's original CIFAR-10 experiment \cite{DBLP:journals/corr/MiikkulainenLMR17} was complete, the resulting best network was trained on all the 50000 training images for 300 epochs. The classification error obtained was 7.3\%, taking 12 epochs to reach 20\% test error and around 120 epochs to converge.
        
        Reproducing such an experiment requires a considerable amount of hardware. Training 100 CNNs for 72 generations and eight training epochs each generation, supposing 30 seconds for each training epoch, would require 1728000 seconds or 480 hours to complete evolution, not considering parallelization efforts. As one of the purposes of this work is to evaluate the usage of CoDeepNEAT in practical use cases for users that might not have access to incredibly potent hardware, the experiments for this work were executed on a smaller scale, similar to the MNIST experiment.
        
        The runs iterated over 40 generations for 6 hours, with populations of 30 modules, 10 blueprints, 10 individuals, and starting with 3 species, running in a setup of 4 cores, 30.5GB memory, no GPU included. Following the same steps as in the MNIST experiment, the elitism rate is set to 20\%, crossover rate is set to 30\%, and the mutation rate is set to 50\%. Training epochs are limited to 4, and the original datasets are down-sampled to 20000 training images and 2000 validation images for early evaluations.
        
\begin{figure}[ht!]
            \centering
                \includegraphics[width=8cm]{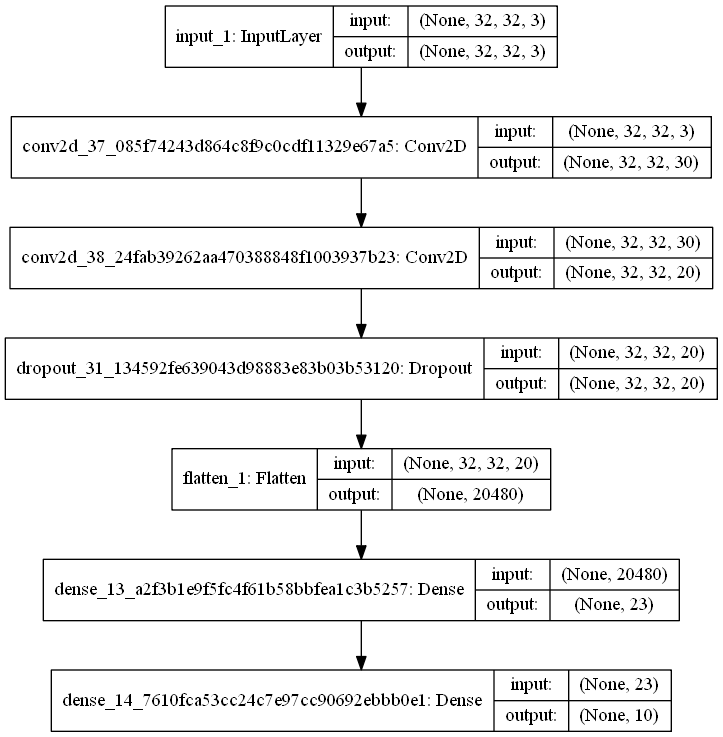}
                \caption{\textbf{Best scoring network for CIFAR-10 at generation $1$.} Smaller network topologies are expected to predominate in early generations.}
                \label{fig:cifar_simple_network}
        \end{figure}

\begin{figure}[ht!]
            \centering
            \includegraphics[width=15cm]{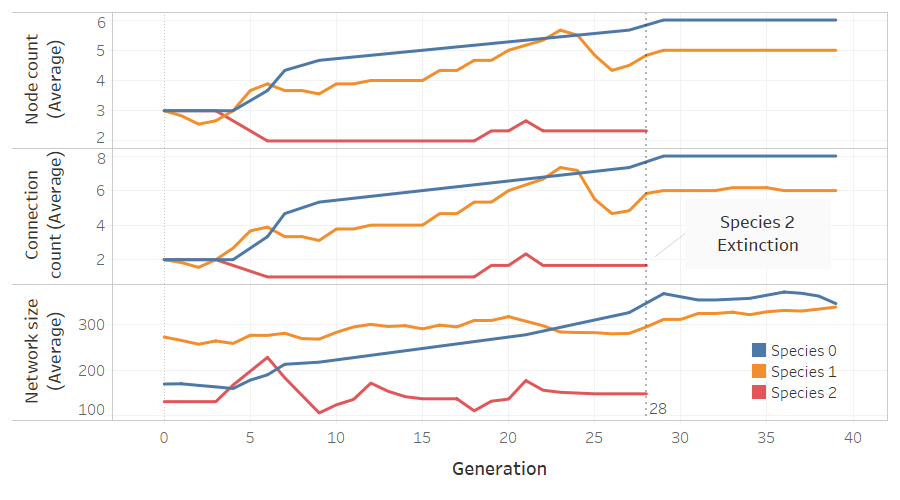}
            \caption{\textbf{Progress of the features of the three species of networks generated for CIFAR-10 over generations.} The representation shows the average of each feature for each species. The different line colors represent different species.}
            \label{fig:cifar_species_evolution}
        \end{figure}

        As expected through the configuration of Table \ref{table:cifar_experiment_hyperp}, initial network structures are small and simple, as the network shown in Figure \ref{fig:cifar_simple_network}, created at the first generation of the experiment. As generations pass, the number of nodes and connections increases or decreases as scores are evaluated. In this specific case, most initial graphs are small structures. Thus they naturally increase over generations. This can be visualized in Figure \ref{fig:cifar_species_evolution}, where the average count of connections, nodes, and the sizes of blueprints increase over generations.
        
\begin{figure}[ht!]
            \centering
                \includegraphics[width=15cm]{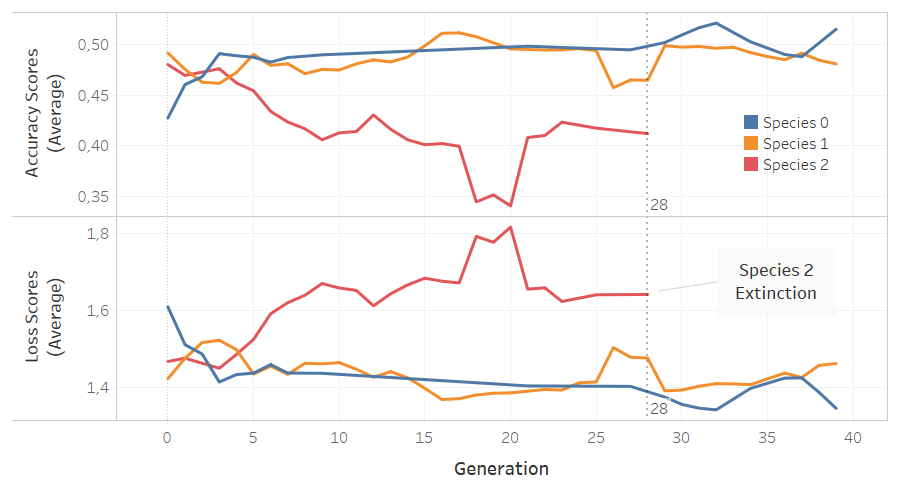}
                \caption{\textbf{Progress of the average accuracy and loss scores of the species over generations for CIFAR-10 dataset.} The different colors represent different species.}
                \label{fig:cifar_species_scores}
        \end{figure}

        The increase in network sizes is expected due to CIFAR-10 being slightly more complex then the use case explored with MNIST, requiring larger structures to differentiate the dataset's classes correctly. Figure \ref{fig:cifar_species_scores} shows the small increase in the accuracy and loss metrics over time as features increase in Figure \ref{fig:cifar_species_evolution}. The improvements in the metrics are small due to the few training epochs used, but this early result has the tendency to impact in greater changes in full training sessions using the complete CIFAR-10 dataset.
        
\begin{figure}[ht!]
            \begin{minipage}[t]{.49\textwidth}
                \begin{subfigure}[b]{\linewidth}
                    \includegraphics[width=7.2cm]{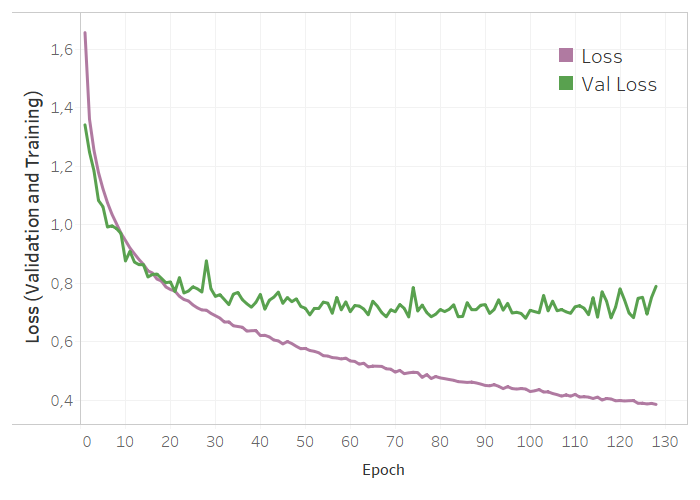}
                    \caption{Loss.}                    
                    \label{fig:cifar_loss}
                \end{subfigure}
            \end{minipage}
            \begin{minipage}[t]{.49\textwidth}
                \begin{subfigure}[b]{\linewidth}
                    \includegraphics[width=7.2cm]{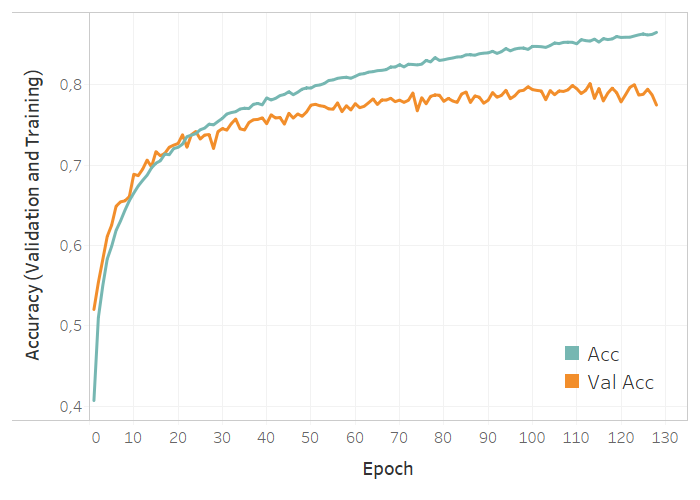}
                   \caption{Accuracy.}
                    \label{fig:cifar_acc}
                \end{subfigure}
            \end{minipage}
            \caption{\textbf{Training and validation metrics for the best network generated for CIFAR-10 after 40 generations.} The network achieved 86.5\% training accuracy and 79.5\% validation accuracy.}
            \label{fig:cifar_metrics}
        \end{figure}

\begin{figure}[ht!]
            \centering
                \includegraphics[width=14cm]{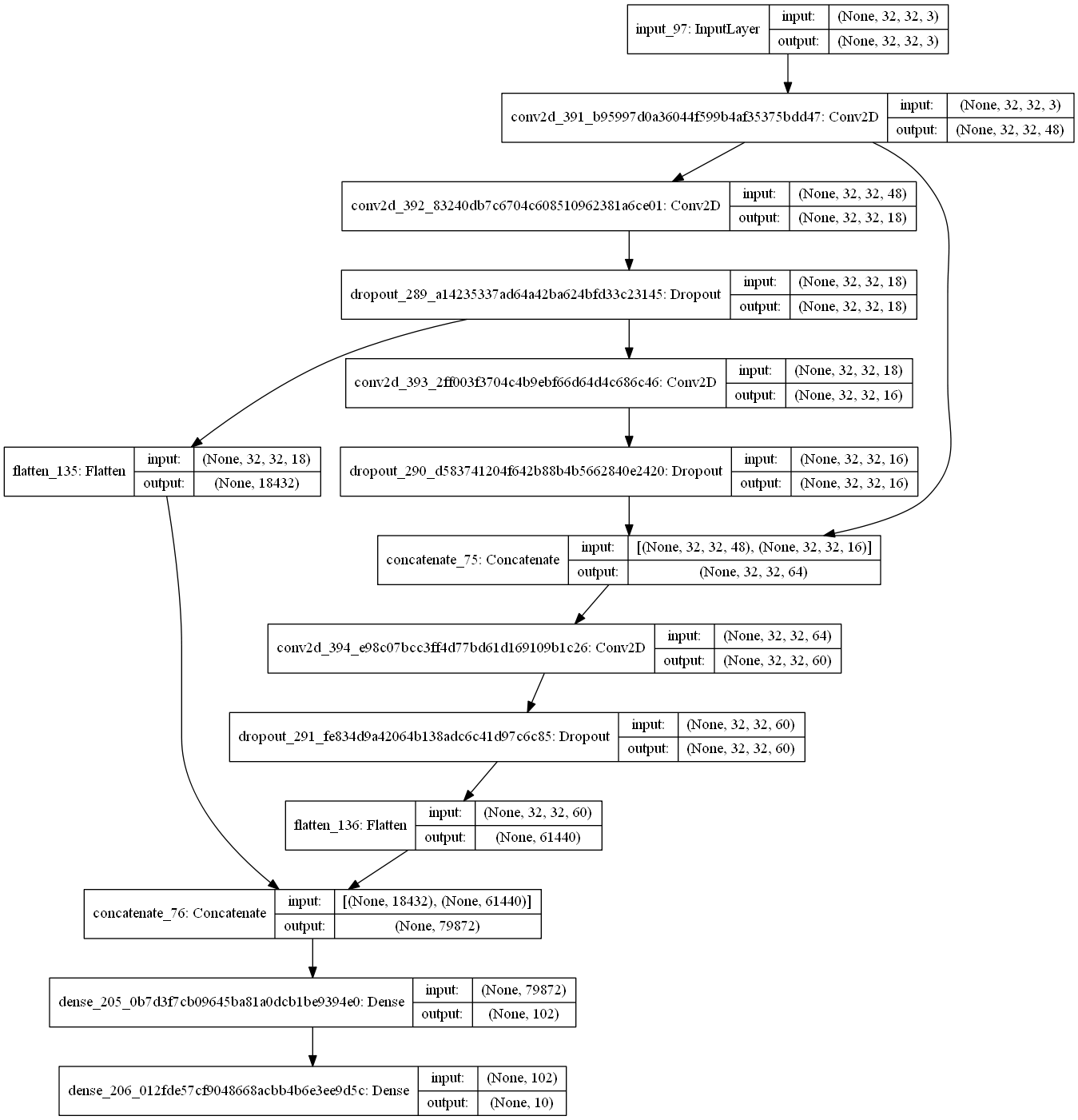}
                \caption{\textbf{Best scoring network for CIFAR-10 at generation $40$.}}
            \label{fig:cifar_final_blueprint}
        \end{figure}

        The best-resulting network from the experiment after 40 generations was obtained in about 10 hours and can be seen in Figure \ref{fig:cifar_final_blueprint}. This network was then trained for 130 epochs but reached a plateau in the validation accuracy metric around the 90th epoch. The achieved training accuracy was of 86.5\% and 79.5\% validation accuracy. Training history for this network can be seen in Figure \ref{fig:cifar_metrics}, depicting loss and accuracy metrics for training and validation datasets.
        
        In comparison to the original CIFAR-10 experiment, the network performed slightly worse, presenting a test accuracy of 77\% (or test error rate of 23\%) as opposed to the 7.3\% error presented by \cite{DBLP:journals/corr/MiikkulainenLMR17}. The convergence of the validation accuracy happened around training epoch 90, converging faster in comparison to the original CIFAR-10, where the convergence occurred around epoch 120. The faster convergence (and subsequent smaller accuracy) are probably associated with the fact that while \cite{DBLP:journals/corr/MiikkulainenLMR17} executed training sessions during evolution using the full CIFAR-10 dataset, this experiment used downsampled versions to reduce the execution time of each generation. At the same time, this experiment considered only four training epochs, while the original used eight training epochs. Evaluating networks with few training epochs usually favors smaller networks that converge faster but do not necessarily achieve optimum accuracy.
        
    \subsection{Discussion}
    
    The results obtained from the experiment show once again that GAs - and specifically CoDeepNEAT - pose as viable solutions to the problems of topology and hyperparameter selection to generate good scoring networks in practical scenarios. Two widely used datasets for machine learning benchmarking and testing, MNIST and CIFAR-10, were used to validate the results and visualize the evolution of solutions over generations.
    
    The proposed implementation returned adequate solutions even with few generations and small population sizes. Still, as research indicates \cite{gasgenerations}, larger populations would probably benefit more from the heuristics used in the genetic algorithm, such as the crossover operator or speciation mechanisms.  The original results from \cite{DBLP:journals/corr/MiikkulainenLMR17} also back this, which evolve much larger populations of solutions throughout almost double the generations and finally generate better scoring networks, with the downside of the more considerable execution time required.
    
    Another point is that studies \cite{Razali2011GeneticAP} indicate that traditional GA operators thrive especially in simple problems where generations can be iterated many times, which is not the case of CoDeepNEAT. The time and hardware requirements for experimentation with large populations and many generations are relatively demanding even for simple use cases like MNIST or CIFAR-10, and are not explored in detail by \cite{DBLP:journals/corr/MiikkulainenLMR17}, making the usage of this type of algorithm very limited to the type of network to be trained and the size of the target dataset.
    
    Deep and complex networks that target significant problems, like high-resolution image recognition \cite{imagenet_cvpr09} or video recognition \cite{ILSVRC15}, for example, may pose as challenging use cases due to the time required for them to execute training sessions even for few epochs. Traditional efforts to improve the efficiency of GAs \cite{CANTUPAZ2000221} may generate minor improvements to execution times. However, the current algorithm would benefit most from approaches that evaluate generated networks using alternative methods rather than exclusively running training sessions for all of them. An example would be methods that generate huge populations but evaluate only certain representatives of each species to elaborate a shared score \cite{934284}, reducing the number of evaluations performed each generation.
    
    Other ideas that could generate benefits to the algorithm are approaches that narrow the search space to more specific topologies, reducing the need to train so many different networks. One recent example would be \cite{Xu_2019_ICCV}, where good results are achieved in reduced GPU time for an object detection use case using the MSCOCO dataset \cite{DBLP:journals/corr/LinMBHPRDZ14} by reducing the search space of the exploration algorithm using specialized topological knowledge on the object detection domain.
    
    In scenarios where computing power is not a problem, CoDeepNEAT is proven to achieve good solutions, as demonstrated by \cite{2019arXiv190206827L}. Then again, the computing power to train thousands of networks by "brute force"  is extremely high and is not commonly accessible for standard users.
    
\section{Conclusion}

    In this work was proposed an open implementation of the CoDeepNEAT algorithm using the popular and highly supported Keras framework. CoDeepNEAT is a powerful neural network topology generation approach based on the neuroevolution of augmenting topologies (NEAT) and the co-evolution of modules. It profits from evolutionary techniques and heuristics to explore the immense search space of possible topological configurations for neural networks, employing specialized genetic algorithm aspects to generate and evaluate solutions to the problems of topology and hyperparameter selection. Even though the algorithm is a known approach, no other accessible and public implementations were available to the general academic community as of the conception of this work.   
    
    The implementation was detailed on how every aspect was designed to fit together in the final version, based on the original algorithm. It was then tested on accessible image datasets and compared to results from the original version considering the differences in environments and experimentation parameters. The results obtained show that acceptable network topologies can be achieved with small population sizes and few generations running in limited hardware environments, even though large runs and large populations generate the best results.
    
    With this implementation complete, possible changes can be proposed to improve the base algorithm, such as different crossover operations, domain-specialized generation rules for topologies, methods for speciation and classification of individuals, and overall population management strategies. The results especially highlight the need to provide better evaluation techniques for the generated neural networks, as it is the most time-consuming activity in the algorithm. Another possibility is improving the network generation procedures to explore specialized topologies with previous domain knowledge, so the necessary network evaluations are narrowed to smaller search spaces.
    
\section*{Data availability statement}    
    
The implementation is available at GitHub (\href{https://github.com/sbcblab/Keras-CoDeepNEAT}{https://github.com/sbcblab/Keras-CoDeepNEAT}) with documentation and examples to reproduce the experiments performed for this work.  

\section*{Acknowledgment}

This work was supported by grants from {FAPERGS} [\textit{16/2551-0000520-6}, 19/2551-0001906-8], {MCT/CNPq} [\textit{311611/2018-4}, Alexander von Humboldt-Stiftung ({AvH}) [\textit{BRA 1190826 HFST} {CAPES-P}] - Germany, and was financed in part by the Coordena\c c\~ao de Aperfei\c coamento de Pessoal de N\'ivel Superior - Brazil ({CAPES}) - Finance Code 001 and CAPES PROBRAL [88881.198766/2018- 01] - Brazil. We gratefully acknowledge the support of the NVIDIA Corporation (hardware grant program). 

The original work was presented by Jonas da Silveira Bohrer in partial fulfillment of the requirements for the degree of Bachelor in Computer Engineering at the Institute of Informatics of the Federal University of Rio Grande do Sul (UFRGS), Brazil, with Marcio Dorn as advisor and Bruno Iochins Grisci as co-advisor. This original text is available at Lume: \href{https://lume.ufrgs.br/}{https://lume.ufrgs.br/}.

\bibliographystyle{unsrt}  
\bibliography{references}  

\begin{thebibliography}{10}

\bibitem{DeJong:2016:ECU:3027779}
Kenneth~A. De~Jong.
\newblock {\em Evolutionary Computation: A Unified Approach}.
\newblock MIT Press, Cambridge, MA, USA, 2016.

\bibitem{Al-Sahaf:2019}
Harith Al-Sahaf, Ying Bi, Qi~Chen, Andrew Lensen, Yi~Mei, Yanan Sun, Binh Tran,
  Bing Xue, and Mengjie Zhang.
\newblock A survey on evolutionary machine learning.
\newblock {\em Journal of the Royal Society of New Zealand}, 49(2):205--228,
  2019.

\bibitem{Floreano2008}
Dario Floreano, Peter D{\"u}rr, and Claudio Mattiussi.
\newblock Neuroevolution: from architectures to learning.
\newblock {\em Evolutionary Intelligence}, 1(1):47--62, Mar 2008.

\bibitem{Back:1996:EAT:229867}
Thomas B\"{a}ck.
\newblock {\em Evolutionary Algorithms in Theory and Practice: Evolution
  Strategies, Evolutionary Programming, Genetic Algorithms}.
\newblock Oxford University Press, Inc., New York, NY, USA, 1996.

\bibitem{Rumelhart1986}
David~E. Rumelhart, Geoffrey~E. Hinton, and Ronald~J. Williams.
\newblock Learning representations by back-propagating errors.
\newblock {\em Nature}, 323(6088):533--536, 1986.

\bibitem{Stanley2019}
Kenneth~O. Stanley, Jeff Clune, Joel Lehman, and Risto Miikkulainen.
\newblock Designing neural networks through neuroevolution.
\newblock {\em Nature Machine Intelligence}, 1(1):24--35, 2019.

\bibitem{Pappa2014}
Gisele~L. Pappa, Gabriela Ochoa, Matthew~R. Hyde, Alex~A. Freitas, John
  Woodward, and Jerry Swan.
\newblock Contrasting meta-learning and hyper-heuristic research: the role of
  evolutionary algorithms.
\newblock {\em Genetic Programming and Evolvable Machines}, 15(1):3--35, Mar
  2014.

\bibitem{gascompetitivereinf}
Felipe~Petroski Such, Vashisht Madhavan, Edoardo Conti, Joel Lehman, Kenneth~O
  Stanley, and Jeff Clune.
\newblock Deep neuroevolution: Genetic algorithms are a competitive alternative
  for training deep neural networks for reinforcement learning.
\newblock {\em arXiv preprint arXiv:1712.06567}, 2017.

\bibitem{salimans2017evolution}
Tim Salimans, Jonathan Ho, Xi~Chen, Szymon Sidor, and Ilya Sutskever.
\newblock Evolution strategies as a scalable alternative to reinforcement
  learning, 2017.

\bibitem{NEAT}
Kenneth~O. Stanley and Risto Miikkulainen.
\newblock Evolving neural networks through augmenting topologies.
\newblock {\em Evolutionary Computation}, 10:99--127, 2001.

\bibitem{PhysRevLett.102.152001}
T.~et~al. Aaltonen.
\newblock Measurement of the top-quark mass with dilepton events selected using
  neuroevolution at cdf.
\newblock {\em Phys. Rev. Lett.}, 102:152001, Apr 2009.

\bibitem{DBLP:journals/corr/ZophL16}
Barret Zoph and Quoc~V. Le.
\newblock Neural architecture search with reinforcement learning.
\newblock {\em CoRR}, abs/1611.01578, 2016.

\bibitem{8790302}
T.~{Watts}, B.~{Xue}, and M.~{Zhang}.
\newblock Blocky net: A new neuroevolution method.
\newblock In {\em 2019 IEEE Congress on Evolutionary Computation (CEC)}, pages
  586--593, June 2019.

\bibitem{pmid30521855}
Bruno~Iochins Grisci, Bruno~C{\'e}sar Feltes, and Marcio Dorn.
\newblock Neuroevolution as a tool for microarray gene expression pattern
  identification in cancer research.
\newblock {\em Journal of biomedical informatics}, 89:122--133, 2019.

\bibitem{cubuk2018autoaugment}
Ekin~D. Cubuk, Barret Zoph, Dandelion Mane, Vijay Vasudevan, and Quoc~V. Le.
\newblock Autoaugment: Learning augmentation policies from data, 2018.

\bibitem{hyperneat}
Kenneth Stanley, David D'Ambrosio, and Jason Gauci.
\newblock A hypercube-based encoding for evolving large-scale neural networks.
\newblock {\em Artificial life}, 15:185--212, 02 2009.

\bibitem{DBLP:journals/corr/MiikkulainenLMR17}
Risto Miikkulainen, Jason~Zhi Liang, Elliot Meyerson, Aditya Rawal, Daniel
  Fink, Olivier Francon, Bala Raju, Hormoz Shahrzad, Arshak Navruzyan, Nigel
  Duffy, and Babak Hodjat.
\newblock Evolving deep neural networks.
\newblock {\em CoRR}, abs/1703.00548, 2017.

\bibitem{chollet2015keras}
Fran\c{c}ois Chollet et~al.
\newblock Keras.
\newblock \url{https://keras.io}, 2015.

\bibitem{tensorflow2015-whitepaper}
Mart\'{\i}n Abadi, Ashish Agarwal, et~al.
\newblock {TensorFlow}: Large-scale machine learning on heterogeneous systems,
  2015.
\newblock Software available from tensorflow.org.

\bibitem{paszke2017automatic}
Adam Paszke, Sam Gross, Soumith Chintala, Gregory Chanan, Edward Yang, Zachary
  DeVito, Zeming Lin, Alban Desmaison, Luca Antiga, and Adam Lerer.
\newblock Automatic differentiation in {PyTorch}.
\newblock In {\em NeurIPS Autodiff Workshop}, 2017.

\bibitem{MNIST}
Y.~{Lecun}, L.~{Bottou}, Y.~{Bengio}, and P.~{Haffner}.
\newblock Gradient-based learning applied to document recognition.
\newblock {\em Proceedings of the IEEE}, 86(11):2278--2324, Nov 1998.

\bibitem{CIFAR-10}
Alex Krizhevsky.
\newblock Learning multiple layers of features from tiny images.
\newblock Technical report, University of Toronto, 2009.

\bibitem{Holland:1992:ANA:531075}
John~H. Holland.
\newblock {\em Adaptation in Natural and Artificial Systems: An Introductory
  Analysis with Applications to Biology, Control and Artificial Intelligence}.
\newblock MIT Press, Cambridge, MA, USA, 1992.

\bibitem{Beasley93anoverview}
David Beasley, David~R. Bull, and Ralph~R. Martin.
\newblock An overview of genetic algorithms : Part 1, fundamentals, 1993.

\bibitem{l.davis1991handbook-of-gen}
L~Davis, editor.
\newblock {\em Handbook of Genetic Algorithms}.
\newblock Van Nostrand Reinhold, 1991.

\bibitem{Unger:1993:GAP:645513.657747}
Ron Unger and John Moult.
\newblock Genetic algorithm for 3d protein folding simulations.
\newblock In {\em Proceedings of the 5th International Conference on Genetic
  Algorithms}, pages 581--588, San Francisco, CA, USA, 1993. Morgan Kaufmann
  Publishers Inc.

\bibitem{Yang1998}
Jihoon Yang and Vasant Honavar.
\newblock {\em Feature Subset Selection Using a Genetic Algorithm}, pages
  117--136.
\newblock Springer US, Boston, MA, 1998.

\bibitem{BORTFELDT2001143}
Andreas Bortfeldt and Hermann Gehring.
\newblock A hybrid genetic algorithm for the container loading problem.
\newblock {\em European Journal of Operational Research}, 131(1):143 -- 161,
  2001.

\bibitem{METAWA201775}
Noura Metawa, M.~Kabir Hassan, and Mohamed Elhoseny.
\newblock Genetic algorithm based model for optimizing bank lending decisions.
\newblock {\em Expert Systems with Applications}, 80:75 -- 82, 2017.

\bibitem{Yuan2017}
Xiaohui Yuan, Mohamed Elhoseny, Hamdy~K. El-Minir, and Alaa~M. Riad.
\newblock A genetic algorithm-based, dynamic clustering method towards improved
  wsn longevity.
\newblock {\em Journal of Network and Systems Management}, 25(1):21--46, Jan
  2017.

\bibitem{Lloyd:2006:LSQ:2263356.2269955}
S.~Lloyd.
\newblock Least squares quantization in pcm.
\newblock {\em IEEE Trans. Inf. Theor.}, 28(2):129--137, September 2006.

\bibitem{DBLP:journals/corr/abs-1205-1117}
T.~Soni Madhulatha.
\newblock An overview on clustering methods.
\newblock {\em CoRR}, abs/1205.1117, 2012.

\bibitem{hastie_09_elements-of.statistical-learning}
Trevor Hastie, Robert Tibshirani, and Jerome Friedman.
\newblock {\em The elements of statistical learning: data mining, inference and
  prediction}.
\newblock Springer, 2 edition, 2009.

\bibitem{934284}
{Hee-Su Kim} and {Sung-Bae Cho}.
\newblock An efficient genetic algorithm with less fitness evaluation by
  clustering.
\newblock In {\em Proceedings of the 2001 Congress on Evolutionary Computation
  (IEEE Cat. No.01TH8546)}, volume~2, pages 887--894 vol. 2, May 2001.

\bibitem{haykin2009neural}
Simon~S. Haykin.
\newblock {\em Neural networks and learning machines}.
\newblock Pearson Education, Upper Saddle River, NJ, third edition, 2009.

\bibitem{4051460}
S.~J. {Mason}.
\newblock Feedback theory-some properties of signal flow graphs.
\newblock {\em Proceedings of the IRE}, 41(9):1144--1156, Sep. 1953.

\bibitem{deng2014deep}
Li~Deng and Dong Yu.
\newblock Deep learning: Methods and applications.
\newblock Technical Report MSR-TR-2014-21, Microsoft Research One Microsoft
  Way, May 2014.

\bibitem{pmlr-v48-amodei16}
Dario Amodei, Sundaram Ananthanarayanan, Zhenyao Zhu, et~al.
\newblock Deep speech 2 : End-to-end speech recognition in english and
  mandarin.
\newblock In Maria~Florina Balcan and Kilian~Q. Weinberger, editors, {\em
  Proceedings of The 33rd International Conference on Machine Learning},
  volume~48 of {\em Proceedings of Machine Learning Research}, pages 173--182,
  New York, New York, USA, 20--22 Jun 2016. PMLR.

\bibitem{Rawat:2017:DCN:3146084.3146086}
Waseem Rawat and Zenghui Wang.
\newblock Deep convolutional neural networks for image classification: A
  comprehensive review.
\newblock {\em Neural Comput.}, 29(9):2352--2449, September 2017.

\bibitem{Goldberg:2017:NNM:3110856}
Yoav Goldberg and Graeme Hirst.
\newblock {\em Neural Network Methods in Natural Language Processing}.
\newblock Morgan \& Claypool Publishers, 2017.

\bibitem{LitjensKBSCGLGS17}
Geert J.~S. Litjens, Thijs Kooi, Babak~Ehteshami Bejnordi, Arnaud
  Arindra~Adiyoso Setio, Francesco Ciompi, Mohsen Ghafoorian, Jeroen A. W.~M.
  van~der Laak, Bram van Ginneken, and Clara~I. Sánchez.
\newblock A survey on deep learning in medical image analysis.
\newblock {\em Medical Image Analysis}, 42:60--88, 2017.

\bibitem{6248110}
D.~{Ciregan}, U.~{Meier}, and J.~{Schmidhuber}.
\newblock Multi-column deep neural networks for image classification.
\newblock In {\em 2012 IEEE Conference on Computer Vision and Pattern
  Recognition}, pages 3642--3649, June 2012.

\bibitem{dsilva:ms06}
Thomas~W. D'Silva.
\newblock Evolving robot arm controllers using the neat neuroevolution method.
\newblock Master's thesis, Department of Electrical and Computer Engineering,
  The University of Texas at Austin, Austin, TX, 2006.

\bibitem{Hastings:2010:GAR:1810136.1810137}
Erin~J. Hastings and Kenneth~O. Stanley.
\newblock Galactic arms race: An experiment in evolving video game content.
\newblock {\em SIGEVOlution}, 4(4):2--10, March 2010.

\bibitem{4983289}
J.~{Clune}, B.~E. {Beckmann}, C.~{Ofria}, and R.~T. {Pennock}.
\newblock Evolving coordinated quadruped gaits with the hyperneat generative
  encoding.
\newblock In {\em 2009 IEEE Congress on Evolutionary Computation}, pages
  2764--2771, May 2009.

\bibitem{hausknecht:tciaig13}
Matthew Hausknecht, Joel Lehman, Risto Miikkulainen, and Peter Stone.
\newblock A neuroevolution approach to general atari game playing.
\newblock {\em IEEE Transactions on Computational Intelligence and AI in
  Games}, 2013.

\bibitem{Fernando:2016:CED:2908812.2908890}
Chrisantha Fernando, Dylan Banarse, Malcolm Reynolds, Frederic Besse, David
  Pfau, Max Jaderberg, Marc Lanctot, and Daan Wierstra.
\newblock Convolution by evolution: Differentiable pattern producing networks.
\newblock In {\em Proceedings of the Genetic and Evolutionary Computation
  Conference 2016}, GECCO '16, pages 109--116, New York, NY, USA, 2016. ACM.

\bibitem{Verbancsics:2011:CCE:2001576.2001776}
Phillip Verbancsics and Kenneth~O. Stanley.
\newblock Constraining connectivity to encourage modularity in hyperneat.
\newblock In {\em Proceedings of the 13th Annual Conference on Genetic and
  Evolutionary Computation}, GECCO '11, pages 1483--1490, New York, NY, USA,
  2011. ACM.

\bibitem{Moriarty1997FormingNN}
David~E. Moriarty and Risto Miikkulainen.
\newblock Forming neural networks through efficient and adaptive coevolution.
\newblock {\em Evolutionary Computation}, 5:373--399, 1997.

\bibitem{Gomez:1999:SNC:1624312.1624411}
Faustino~J. Gomez and Risto Miikkulainen.
\newblock Solving non-markovian control tasks with neuroevolution.
\newblock In {\em Proceedings of the 16th International Joint Conference on
  Artificial Intelligence - Volume 2}, IJCAI'99, pages 1356--1361, San
  Francisco, CA, USA, 1999. Morgan Kaufmann Publishers Inc.

\bibitem{gomez:jmlr08}
Faustino Gomez, Juergen Schmidhuber, and Risto Miikkulainen.
\newblock Accelerated neural evolution through cooperatively coevolved
  synapses.
\newblock {\em Journal of Machine Learning Research}, pages 937--965, 2008.

\bibitem{DBLP:journals/corr/ChenFLVGDZ15}
Xinlei Chen, Hao Fang, Tsung{-}Yi Lin, Ramakrishna Vedantam, Saurabh Gupta,
  Piotr Doll{\'{a}}r, and C.~Lawrence Zitnick.
\newblock Microsoft {COCO} captions: Data collection and evaluation server.
\newblock {\em CoRR}, abs/1504.00325, 2015.

\bibitem{2019arXiv190206827L}
Jason {Liang}, Elliot {Meyerson}, Babak {Hodjat}, Dan {Fink}, Karl {Mutch}, and
  Risto {Miikkulainen}.
\newblock {Evolutionary Neural AutoML for Deep Learning}.
\newblock {\em arXiv e-prints}, page arXiv:1902.06827, Feb 2019.

\bibitem{rajpurkar2017chexnet}
Pranav Rajpurkar, Jeremy Irvin, Kaylie Zhu, Brandon Yang, Hershel Mehta, Tony
  Duan, Daisy Ding, Aarti Bagul, Curtis Langlotz, Katie Shpanskaya, Matthew~P.
  Lungren, and Andrew~Y. Ng.
\newblock Chexnet: Radiologist-level pneumonia detection on chest x-rays with
  deep learning, 2017.

\bibitem{SciPyProceedings_11Networkx}
Aric~A. Hagberg, Daniel~A. Schult, and Pieter~J. Swart.
\newblock Exploring network structure, dynamics, and function using networkx.
\newblock In Ga\"el Varoquaux, Travis Vaught, and Jarrod Millman, editors, {\em
  Proceedings of the 7th Python in Science Conference}, pages 11 -- 15,
  Pasadena, CA USA, 2008.

\bibitem{scikit-learn}
F.~Pedregosa, G.~Varoquaux, A.~Gramfort, V.~Michel, B.~Thirion, O.~Grisel,
  M.~Blondel, P.~Prettenhofer, R.~Weiss, V.~Dubourg, J.~Vanderplas, A.~Passos,
  D.~Cournapeau, M.~Brucher, M.~Perrot, and E.~Duchesnay.
\newblock Scikit-learn: Machine learning in {P}ython.
\newblock {\em Journal of Machine Learning Research}, 12:2825--2830, 2011.

\bibitem{Holland:1975}
John~H. Holland.
\newblock {\em Adaptation in Natural and Artificial Systems}.
\newblock University of Michigan Press, Ann Arbor, MI, 1975.
\newblock second edition, 1992.

\bibitem{DBLP:journals/jece/MattioliCCNL19}
Fernando Mattioli, Daniel Caetano, Alexandre Cardoso, Eduardo L.~M. Naves, and
  Edgard Lamounier.
\newblock An experiment on the use of genetic algorithms for topology selection
  in deep learning.
\newblock {\em J. Electrical and Computer Engineering},
  2019:3217542:1--3217542:12, 2019.

\bibitem{DBLP:journals/corr/SimonyanZ14a}
Karen Simonyan and Andrew Zisserman.
\newblock Very deep convolutional networks for large-scale image recognition.
\newblock {\em CoRR}, abs/1409.1556, 2014.

\bibitem{WittenFrankHall11}
Ian~H. Witten, Eibe Frank, and Mark~A. Hall.
\newblock {\em Data Mining: Practical Machine Learning Tools and Techniques}.
\newblock Morgan Kaufmann Series in Data Management Systems. Morgan Kaufmann,
  Amsterdam, 3 edition, 2011.

\bibitem{gasgenerations}
Dana Vrajitoru.
\newblock Large population or many generations for genetic algorithms?
  implications in information retrieval.
\newblock In {\em Soft Computing in Information Retrieval}, pages 199--222.
  Springer, 2000.

\bibitem{Razali2011GeneticAP}
Noraini~Mohd Razali, John Geraghty, et~al.
\newblock Genetic algorithm performance with different selection strategies in
  solving tsp.
\newblock In {\em Proceedings of the world congress on engineering}, volume~2,
  pages 1--6. International Association of Engineers Hong Kong, 2011.

\bibitem{imagenet_cvpr09}
J.~Deng, W.~Dong, R.~Socher, L.-J. Li, K.~Li, and L.~Fei-Fei.
\newblock {ImageNet: A Large-Scale Hierarchical Image Database}.
\newblock In {\em CVPR09}, 2009.

\bibitem{ILSVRC15}
Olga Russakovsky, Jia Deng, Hao Su, Jonathan Krause, Sanjeev Satheesh, Sean Ma,
  Zhiheng Huang, Andrej Karpathy, Aditya Khosla, Michael Bernstein,
  Alexander~C. Berg, and Li~Fei-Fei.
\newblock {ImageNet Large Scale Visual Recognition Challenge}.
\newblock {\em International Journal of Computer Vision (IJCV)},
  115(3):211--252, 2015.

\bibitem{CANTUPAZ2000221}
Erick Cantú-Paz and David~E. Goldberg.
\newblock Efficient parallel genetic algorithms: theory and practice.
\newblock {\em Computer Methods in Applied Mechanics and Engineering},
  186(2):221 -- 238, 2000.

\bibitem{Xu_2019_ICCV}
Hang Xu, Lewei Yao, Wei Zhang, Xiaodan Liang, and Zhenguo Li.
\newblock Auto-fpn: Automatic network architecture adaptation for object
  detection beyond classification.
\newblock In {\em The IEEE International Conference on Computer Vision (ICCV)},
  October 2019.

\bibitem{DBLP:journals/corr/LinMBHPRDZ14}
Tsung{-}Yi Lin, Michael Maire, Serge~J. Belongie, Lubomir~D. Bourdev, Ross~B.
  Girshick, James Hays, Pietro Perona, Deva Ramanan, Piotr Doll{\'{a}}r, and
  C.~Lawrence Zitnick.
\newblock Microsoft {COCO:} common objects in context.
\newblock {\em CoRR}, abs/1405.0312, 2014.

\end{thebibliography}

\end{document}